%% file: main.tex
\documentclass{article}

\usepackage[final]{neurips_2023}
\usepackage{natbib}
\setcitestyle{authoryear}

\usepackage[utf8]{inputenc} 
\usepackage[T1]{fontenc}    
\usepackage{hyperref}       
\usepackage{url}            
\usepackage{booktabs}       
\usepackage{amsfonts}       
\usepackage{nicefrac}       
\usepackage{microtype}      
\usepackage{xcolor}         

\definecolor{bluecite}{HTML}{0875b7}
\hypersetup{
  colorlinks=true,
  citecolor=bluecite,
  linkcolor=magenta,
  linktoc=section,
}

\usepackage{graphicx}
\usepackage{wrapfig}
\usepackage{caption}
\usepackage{subcaption}
\usepackage{amsmath,amssymb}
\usepackage{array,multirow}
\usepackage{xspace}
\usepackage{tabularx}
\usepackage{float}
\usepackage{siunitx}
\usepackage[capitalise]{cleveref}

\usepackage{minitoc}
\usepackage{tocloft}

\usepackage{algorithm}
\usepackage[noend]{algorithmic}
\newcommand{\FuncSty}[1]{\textnormal{\texttt{#1}}\unskip}
\DeclareMathOperator*{\argmax}{arg\,max}

\newcommand{\poppy}{{Poppy }\xspace}
\newcommand{\compass}{{COMPASS}\xspace}

 \newcommand{\del}[1]{}
\newcommand{\ins}[1]{#1}
 \newcommand{\rep}[2]{\ins{#2}}

\title{Combinatorial Optimization with Policy Adaptation using Latent Space Search}

\author{%
  \textbf{Felix Chalumeau$^*$} \\
  InstaDeep \\
  \texttt{f.chalumeau@instadeep.com} \\
  \and
  \textbf{Shikha Surana$^*$} \\
  InstaDeep \\
  \texttt{s.surana@instadeep.com} \\
  \and
  \textbf{Clément Bonnet} \\
  InstaDeep \\
  \texttt{c.bonnet@instadeep.com} \\
  \and
  \textbf{Nathan Grinsztajn} \\
  InstaDeep \\
  \texttt{n.grinsztajn@instadeep.com} \\
  \and
  \textbf{Arnu Pretorius} \\
  InstaDeep \\
  \texttt{a.pretorius@instadeep.com} \\
  \and
  \textbf{Alexandre Laterre} \\
  InstaDeep \\
  \texttt{a.laterre@instadeep.com} \\
  \and
  \textbf{Thomas D. Barrett} \\
  InstaDeep \\
  \texttt{t.barrett@instadeep.com} \\
}

\begin{document}

\maketitle

\def\thefootnote{*}\footnotetext{Equal contribution}\def\thefootnote{\arabic{footnote}}

\doparttoc
\faketableofcontents

\input{sections/0-abstract}
\input{sections/1-introduction}
\input{sections/2-related_work}
\input{sections/3-method}

\input{sections/4-experiments}
\input{sections/5-conclusion}

\section*{Acknowledgements}
Research supported with Cloud TPUs from Google's TPU Research Cloud (TRC). We thank anonymous reviewers for comments and helpful discussions that helped improve the paper.

\bibliographystyle{abbrvnat}
\bibliography{references}

\newpage
\appendix
\addcontentsline{toc}{section}{Appendix}
\part{Appendix}

{\hypersetup{hidelinks}
\parttoc
}

\newpage

\input{sections/99-appendix}

\end{document}

%% file: sections/0-abstract.tex
\begin{abstract}

Combinatorial Optimization underpins many real-world applications and yet, designing performant algorithms to solve these complex, typically NP-hard, problems remains a significant research challenge. Reinforcement Learning (RL) provides a versatile framework for designing heuristics across a broad spectrum of problem domains. However, despite notable progress, RL has not yet supplanted industrial solvers as the go-to solution. Current approaches emphasize pre-training heuristics that construct solutions but often rely on search procedures with limited variance, such as stochastically sampling numerous solutions from a single policy or employing computationally expensive fine-tuning of the policy on individual problem instances. Building on the intuition that performant search at inference time should be anticipated during pre-training, we propose COMPASS, a novel RL approach that parameterizes a distribution of diverse and specialized policies conditioned on a continuous latent space. We evaluate COMPASS across three canonical problems - Travelling Salesman, Capacitated Vehicle Routing, and Job-Shop Scheduling - and demonstrate that our search strategy (i) outperforms state-of-the-art approaches on 11 standard benchmarking tasks and (ii) generalizes better, surpassing all other approaches on a set of 18 procedurally transformed instance distributions.

\end{abstract}

%% file: sections/1-introduction.tex
\section{Introduction}

Combinatorial Optimization (CO) has a wide range of real-world applications, from transportation~\citep{contardo2012balancing} and logistics~\citep{laterre2018ranked}, to energy~\citep{FROGER2016695}. Solving a CO problem consists of finding an ordering, labelling or subset of elements from a finite, discrete set that maximizes (or minimizes) a given objective function. As the number of feasible solutions typically grows exponentially with the problem size, CO problems are challenging (often NP-hard) to solve.  As such, significant work goes into designing problem-specific heuristic approaches that, whilst not guaranteeing the optimal answer, can often work well in practice. Reinforcement Learning (RL) offers a domain-agnostic framework to learn heuristics and has been successfully applied across a range of CO tasks~\citep{pointernetworks, Deudon2018, MAZYAVKINA2021105400}.

Concretely, leading RL methods typically train a policy to incrementally \textit{construct} a solution one element at a time. However, whilst most efforts have focused on improving the one-shot quality of these construction heuristics~\citep{Kool2019, POMO, poppy}, it intuitively appears impractical to reliably produce the optimal solution to NP-hard problems within a single construction attempt. Consequently, competitive performance has to rely on combining a pre-trained policy with an additional search procedure. Nevertheless, this crucial aspect is often implemented using simple procedures such as stochastic sampling~\citep{Kool2019, POMO, poppy}, beam search~\citep{steinbiss1994improvements} or Monte Carlo Tree Search (MCTS)~\citep{browne2012survey}. An alternative approach, representing the current state-of-the-art for search-based RL~\citep{Bello16, hottung2022efficient}, is to actively re-train the heuristic on each new problem instance; however, this comes with clear computational and practical limitations. Strikingly, neither of these approaches pre-trains the policy in a way that could enable a fast and efficient inference time search: rather current approaches typically completely decouple both. The absence of an efficient search strategy is even more detrimental when the test instances are out of the distribution (OOD) used to train the policy, as this may cause a large difference between the learned policy and the policy leading to the optimal solution.

In this work, we aim to overcome the current limitations of search strategies used in RL when applied to CO problems. Our approach is to learn a latent space of diverse policies that can be explored at inference time in order to find the best-performing strategies for a given instance. This updates the current paradigm by enabling sampling from a policy space at inference time rather than constantly sampling the same policy (or set of policies) with stochasticity. We introduce \textbf{\compass} -- \textbf{COM}binatorial optimization with \textbf{P}olicy \textbf{A}daptation using Latent \textbf{S}pace \textbf{S}earch. \compass effectively creates an infinite set of diverse solvers by using a single conditioned policy and sampling the conditions from a continuous latent space. The training process encourages subareas of the latent space to specialize to sub-distributions of instances and this diversity is used at inference time to solve newly encountered instances.

We evaluate \compass on three popular CO problems: Travelling Salesman Problem (TSP), Capacitated Vehicle Routing Problem (CVRP) and Job-Shop Scheduling Problem (JSSP). After training on a distribution of fixed-sized instances for each problem, we evaluate our method on both in- and out-of-distribution test sets.  We find that simple search strategies requiring no re-training provide both rapid and sustained improvement of the instance-specific policy, with \compass establishing a new state-of-the-art across all problems in this setting. Thanks to the diversity provided by its latent space, COMPASS achieves high performance even without a search budget and achieves comparable or better results than current leading few-shot methods.

Concretely, our work makes the following contributions: \textbf{(i)} We introduce \compass which leverages a latent space of diverse and specialized policies to effectively solve CO problems. \textbf{(ii)} We show that \compass allows for the efficient adaptation of instance-specific policies without re-training or sacrificing zero-shot performance. \textbf{(iii)} Experimentally, our approach is found to represent a new state-of-the-art for RL-based CO methods across all our considered problem types, achieving superior performance on all 29 tasks. \textbf{(iv)} We release fast and performant implementations of our method and its main competitors, written in JAX.  We also provide all of our test sets including our procedurally transformed problem instances for easier comparison in future work.

%% file: sections/2-related_work.tex
\section{Related work}
\label{related_work}

\paragraph{Construction methods for CO}

Construction approaches in RL for CO incrementally build a solution by selecting one element at a time. After \citet{Hopfield85} first applied neural networks to TSP, \citet{Bello16} extended these efforts by proposing to learn heuristics with RL using a Pointer Network \citep{pointernetworks} combined with an actor-critic framework. This approach was extended by \citet{Deudon2018} who added an attention-based city encoder, which was subsequently further extended by \citet{Kool2019} to use a general transformer architecture \citep{Attention_all}. The transformer has since become the standard model for a range of CO problems and is also used in this work. \citet{kim2022symnco} builds on \citet{Kool2019} by leveraging symmetries of routing problems during training. Even though the majority of these construction approaches have focused on routing problems, numerous works have also tackled other classes of CO problems, especially on graphs, like Maximum Cut \citep{Dai17, Tom2019}, or Job Shop Scheduling Problem (JSSP), for which \citet{Zhang2020} proposed a Graph Neural Network (GNN) approach. A broader scope of (non-construction) approaches can be found in Appendix~\ref{appendix:extended-related-work}.

\paragraph{Improving solutions at inference time}
As it is unlikely that the first solution generated by a construction heuristic is optimal, a popular approach consists in sampling various trajectories during inference for the same CO problem. POMO~\citep{POMO} uses one policy rolled out on several versions of the same problem, while considering different starting points or symmetries, to create diverse trajectories and select the best one. \citet{choo2022simulationguided} proposes an efficient search guided by simulations, but cannot take advantage of a large inference budget by itself. EAS~\citep{hottung2022efficient} adds on POMO by fine-tuning a subset of the model parameters at inference time using gradient descent. However, the new solutions are biased toward the underlying pre-trained policy and can easily be stuck in local optima. Instead, MDAM~\citep{XinSCZ21} and Poppy~\citep{poppy} employ a population of agents, all of which are simultaneously rolled out at inference time. MDAM trains these policies to select different initial actions, whereas Poppy utilizes a loss function designed to specialize each policy on specific subsets of the problem distribution. Despite demonstrating promising performance, these approaches are constrained by the number of policies used during training, which remains fixed. Such a limitation quickly diminishes the benefits of additional solution candidates sampled from the population. Our method \compass uses the same loss as Poppy, but, unlike their approach and that of MDAM, \compass is not bound to a specific number of specialized policies. Moreover, its latent space makes it possible to add additional search mechanisms over the policy space, ensuring better solutions over time. CVAE-Opt~\citep{Hottung2021}, akin to our method, uses a latent space for solving routing problems, however, it has several differences. First, \compass is trained end-to-end with RL, hence does not necessitate pre-solved instances. Second, CVAE-Opt requires training an additional recurrent encoder for (instance, solution) pairs, whereas \compass uses the latent space to encode a distribution of complementary policies and can be easily applied to pre-train models. Overall, \compass significantly outperforms CVAE-Opt while having shorter runtime.

%% file: sections/3-method.tex
\section{Methods}
\label{method}

\subsection{Preliminaries}

\paragraph{Formulation} 
The goal of a CO problem is to find the optimal labeling of a set of discrete variables that satisfies the problem's constraints. In RL, a CO problem can be formulated as a Markov Decision Process (MDP) defined by $M = (S, A, R, T, \gamma, H)$. This includes the state space $S$ with states \(s_i \in S\), action space $A$ with actions \(a_i \in A\), reward function \(R: S \times A \rightarrow R\), transition function \(T(s_{i+1}|s_i, a_i)\), discount factor \(\gamma \in [0, 1]\), and horizon \(H\) which denotes the episode duration. The state of a problem instance is represented as the (partial) trajectory or set of actions taken in the instance, and the next state \(s_{t+1}\) is determined by applying the chosen action \(a_t\) to the current state \(s_t\). 
An agent is introduced in the MDP to interact with the CO problem and find solutions by learning a policy $\pi: S \rightarrow A$. The policy is trained to maximize the expected sum of discounted rewards to find the optimal solution, and this is formalized as the following learning objective: \(\pi^* = \underset{\pi}{\mathrm{argmax}}\ \mathbb{E}[\sum_{t=0}^H \gamma^tR(s_t, a_t)]\).

\subsection{\compass}
\label{sec:method}

Recall our intuition that no single policy will reliably be able to solve all instances of an NP-hard CO problem in a single inference pass. Two primary approaches to address this are the inclusion of inference time search and the deployment of a diverse set of policies to increase the chance of a near-optimal strategy being deployed.  This work aims to unify and extend these approaches by training an infinitely large set of diverse and specialized policies that can subsequently be searched at inference time.

\begin{figure}[t]
    \centering
    \includegraphics[width=0.9\linewidth]{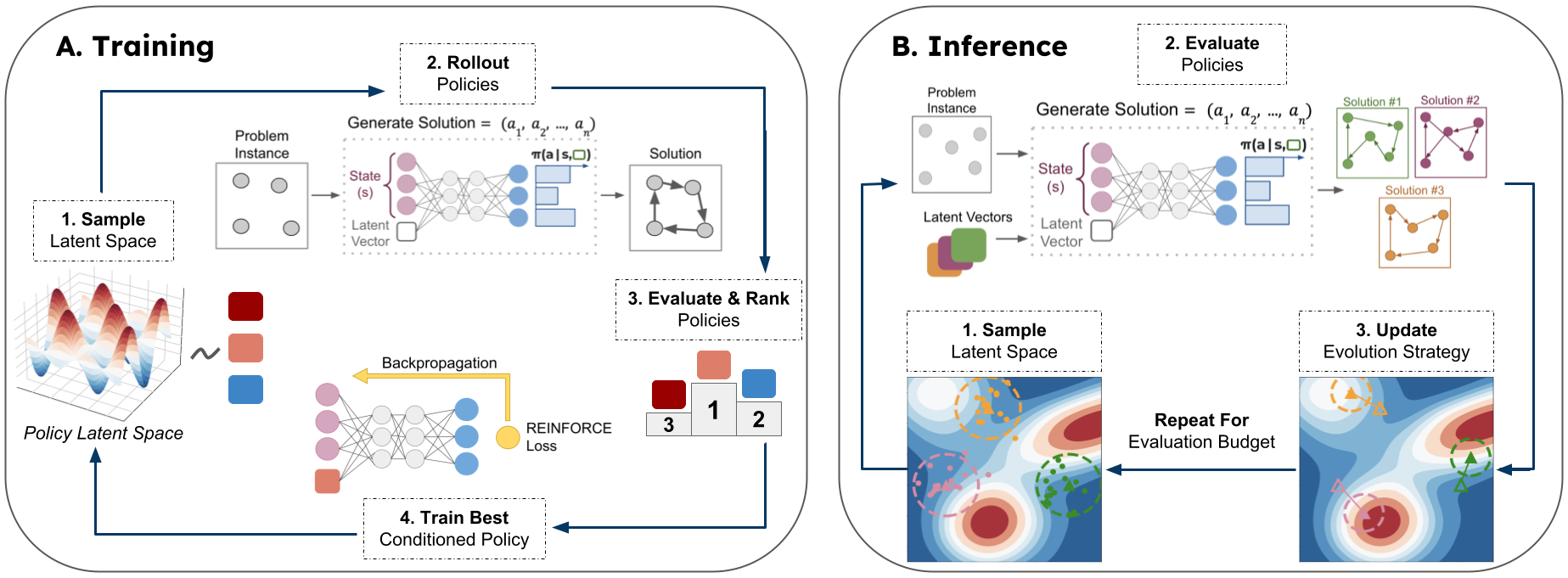}
    \caption{Our method \compass is composed of the following two phases. A. Training - the latent space is sampled to generate vectors that the policy can condition upon. The conditioned policies are then evaluated and only the best one is trained to create specialization within the latent space. B. Inference - at inference time the latent space is searched through an evolution strategy to exploit regions with high-performing policies for each instance.}
    \label{fig:compass}
\end{figure}

To achieve this, we propose that a single set of policy parameters condition not just on the current observation, but also on samples drawn from a continuous latent space.  The training objective then encourages this latent space of policies to be diverse (generate a wide range of behaviors) and specialized (these behaviors are optimized for different types of problem instances from the training distribution).
This latent space can then be efficiently searched during inference to find the most performant policy for a given problem instance.
In this section, we describe in detail the realization of this approach, which we call \compass (\textbf{COM}binatorial optimization with \textbf{P}olicy \textbf{A}daptation using Latent \textbf{S}pace \textbf{S}earch). In \cref{fig:compass}, we provide an illustrated overview of \compass.

Our approach offers several key advantages over traditional techniques. Compared to methods that directly train multiple, uniquely parameterized policies \citep{XinSCZ21, poppy}, training a single conditional policy can, in principle, provide a continuous distribution of an infinite number of policies. Moreover, our approach mitigates the significant training and memory overheads associated with training a population of agents. Compared to methods that rely on brute-force sampling \citep{Kool2019, POMO, poppy} or expensive fine-tuning \citep{hottung2022efficient},  our training process produces a structured latent space (where similar policies are found near to each other) that permits principled search during inference.

\paragraph{Latent space}
The latent space defines the set of policies that our model can condition itself upon. Importantly, we do not learn the distribution of this space, but rather select a prior distribution over the space from which we sample during training. In practice, we use a latent space with 16 dimensions bounded between -1 and 1, and use a uniform sampling prior.

\paragraph{Architecture}
\compass is agnostic to the network architecture used, so long as the resulting policy is, in some way, conditioned on the vector sampled from the latent space.  This can be achieved in numerous ways, from directly concatenating the vector to the input observation to conditioning keys, queries, and values in the self-attention models commonly used for CO. We refer to \cref{app:network-details} for further details about the architectures used in this work and how the latent vector is used to condition them. Whilst it is possible to train \compass from scratch, we found that it was simple and efficient to adapt pre-trained single-policy models. To do this, we zero-initialize any additional weights corresponding to the sample latent vector such that it has no impact at the start of training. In practice, we adapt a single-agent architecture designed for few-shot inference in all of our problem settings; POMO~\citep{POMO} for TSP and CVRP, and a similar architecture taken from Jumanji~\citep{jumanji2023github} for JSSP (we also considered the current SOTA model L2D~\citep{Zhang2020}, however, we found that the model from Jumanji already outperformed this approach). Full network details can be found in Appendices \ref{app:tsp-crvp-network-details} (TSP \& CVRP) and \ref{sec:appendix-jssp-architecture} (JSSP).

\paragraph{Training}
The training procedure aims to specialize subareas of the latent space to sub-distributions of problems by training the policy solely on latent vectors that achieve the best performance on a given problem. At each training step, we uniformly sample a set of $N$ vectors from the latent space and condition the policy on each vector resulting in $N$ conditioned policies. After evaluating each policy on the problem instance, we train the best policy (i.e.\ the policy conditioned on the best-performing latent vector) on the instance. The model is updated using the gradient of our objective as given by

\begin{equation}
\label{eq:loss_func}    
\begin{aligned}
    \nabla_\theta J_{\text{compass}} = \mathbb{E}_{\rho \sim \mathcal{D}} \mathbb{E}_{z_1, ..., z_N \sim \mathcal{P}_z} \mathbb{E}_{\tau_i \sim \pi_\theta(\cdot |z_i)} [&\nabla_\theta \log \pi_\theta(\tau_{i^\star} | z_{i^\star})(R_{i^\star} - \mathcal{B}_{\rho, \theta})], \\
\end{aligned}
\end{equation}

where $\mathcal{D}$ is the data distribution, $\mathcal{P}_z$ the latent space, $z_i$ a latent vector, $\pi_\theta$ the conditioned policy, $\tau_i$ the trajectory generated by policy $\pi_\theta$ conditioned on vector $z_i$ and has the corresponding reward $R_i$, $i^\star$ is the index of the best performing latent vector (in the sampled set) and is expressed as $i^\star = \arg \max_{i \in [1, N]} R(\tau_i)$, and lastly, $\mathcal{B}_{\rho, \theta}$ is the baseline, inspired by~\citet{POMO}. Full details of the algorithmic procedure can be found in~\cref{sec:appendix-training-process}. Notably, our work is the first to create a specialized and diverse set of policies represented by a continuous latent space by only training the best-performing vector for each problem instance. 

A key training hyperparameter is the number of condition vectors sampled during evaluation. More conditioned policies results in an increased certainty that the best-performing vector in the sampled set of conditions is the best-performing vector in the latent space. Therefore, increasing the number of sampled conditions increases the likelihood of training the true best latent vector for the given problem instance, rather than a potentially suboptimal vector. More details (including training times and environment steps) are reported in~\cref{sec:appendix-training-process}.

\paragraph{Inference-time search} 
Given the latent space of diverse, specialized policies obtained by training \compass, at inference time, we apply a principled search procedure to find the most performant strategies. Our desired properties for a search procedure are that it should be simple, capable of rapid adaptation and robust to local optima.  As such, evolutionary strategies are an appropriate approach. Specifically, we use Covariance Matrix Adaptation (CMA-ES, \citep{hansen2001cmaes}). CMA-ES uses a multivariate normal distribution to sample vectors and iteratively updates the distribution's mean to increase the expected performance of sampled vectors (i.e.\ the quality of the solution found by the policy corresponding to each vector). The covariance is also adapted over time, either for exploration (high values, broad sampling) or exploitation (small values, focused sampling).

For a given problem instance, there may be multiple high-performance policies, therefore we use several independent CMA-ES components in parallel. To ensure that those components explore distinct areas of the space (or at least, take different paths), we compute a Voronoi Tesselation~\citep{du1999cvt} of the latent space and use the corresponding centroids to initialize the means of the CMA-ES components. This method proves to be robust, easy to tune, and fast, and requires low memory and computation budget, making it the perfect candidate for efficient adaptation at inference time. In our experimental section (\ref{sec:exp-search-strategies}), we present an analysis of our latent space and how it is explored by CMA-ES. Details and considered alternatives can be found in~\cref{sec:appendix-search-strategy}.

%% file: sections/4-experiments.tex
\section{Experiments}
\label{experiments}

\begin{figure}[b]
    \centering
    \begin{minipage}[c]{0.45\textwidth}
    \centering
    \includegraphics[width=0.95\linewidth]{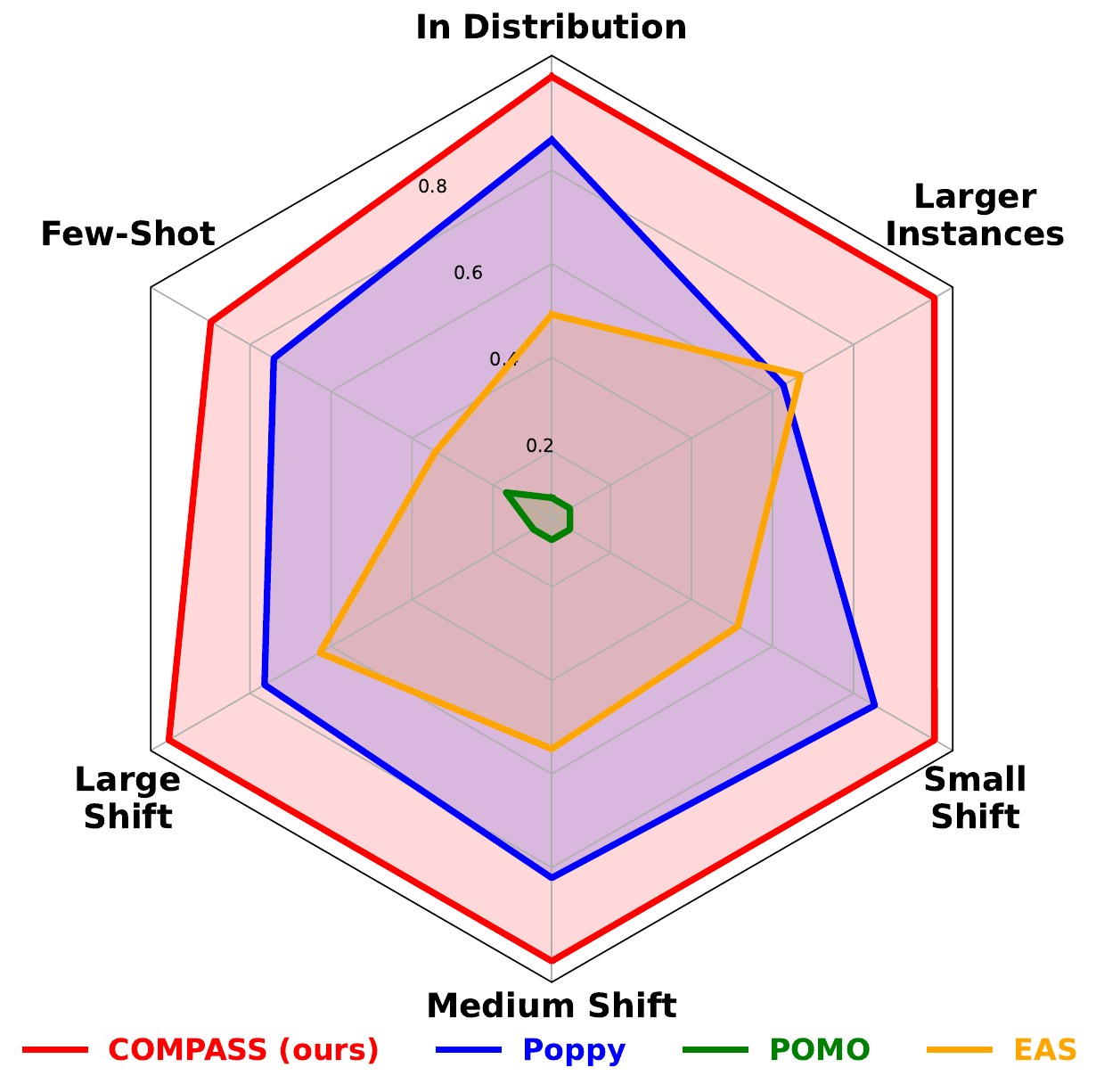}
    \end{minipage}
    \hfill
    \begin{minipage}[l]{0.5\textwidth}
        \paragraph{In Distribution} Instances drawn from the training distribution (see first cols.\ in \cref{tab:tsp_size,tab:cvrp_size,tab:jssp_size}).
        \vspace{3pt}
        \paragraph{Larger Instances} Instance with a larger size than the training data. See \emph{generalization} in \cref{tab:tsp_size,tab:cvrp_size,tab:jssp_size}.
        \vspace{3pt}
        \paragraph{Small/Medium/Large Shift} Instances that have been procedurally mutated to be out of the training distribution. Detailed results in \cref{fig:distrib-shift-generalisation}.
        \vspace{3pt}
        \paragraph{Few-Shot} The standard benchmark with a lower budget (typically 10\% of the usual). Numerical results in Appendix~\ref{sec:appendix-fast-rl}.
    \end{minipage}
\caption{Performance of COMPASS and the main baselines aggregated across several tasks over three problems (TSP, CVRP, and JSSP). For each task (problem type, instance size, mutation power), we normalize values between 0 and 1 (corresponding to the worst and best performing RL method, respectively). Hence, all tasks have the same impact on the aggregated metrics. \compass surpasses the baselines on all of them, showing its versatility for all types of tasks and in particular, its generalization capacity.}    
\label{fig:radar}
\end{figure}

We evaluate our method on three problems -- Travelling Salesman (TSP), Capacitated Vehicle Routing (CVRP), and Job Shop Scheduling (JSSP) -- widely used to assess RL-based methods for CO \citep{Deudon2018, Kool2019, poppy, hottung2022efficient}. In Section~\ref{sec:larger_instances}, we evaluate \compass in the standard setting used by other methods from the literature and report results on each problem type. In Section~\ref{sec:mutated_sets}, we assess the robustness of methods by evaluating them on instances of TSP and CVRP that are procedurally transformed using the approach developed by~\citet{diverse_tsp_instances}. In Section~\ref{sec:exp-search-strategies}, we analyze the methods' search strategies; in particular, we provide insights about \compass' latent space and how it is navigated by CMA-ES at inference time. Figure~\ref{fig:radar} provides a radar plot overview of our aggregated experimental results across six performance categories of interest: (1) in distribution instances, OOD instances with different levels of distribution shift (2) small, (3) medium and (4) large, (5) large instance sizes as well as (6) few-shot performance. Our results highlight the strengths and weaknesses of each approach and in particular, the versatility and superiority of \compass.

\paragraph{Baselines} 
We compare \compass to a suite of leading RL methods and industrial solvers. Across all problems we provide baselines for EAS~\citep{hottung2022efficient}; the current SOTA active-search RL method that fine-tunes the policy on each problem instance, and Poppy~\citep{poppy}; the current SOTA active-search RL method that stochastically samples from a fixed population of pre-trained solvers. For routing problems (TSP and CVRP), we also provide results for POMO~\citep{POMO}; the leading single-agent, one-shot architecture on which EAS and Poppy are built, and LKH~\citep{lkh3}; a leading industrial solver. We also report results of TSP-specific industrial solver Concorde~\citep{concorde}.
For JSSP, we provide results for L2D~\citep{Zhang2020}; the leading single-agent, one-shot architecture. We also provide results for the attention-based model proposed in Jumanji~\citep{jumanji2023github} that proved to outperform L2D. Finally, we report the results of Google OR-Tools \citep{perronor}; the reference industrial solver for JSSP.

\paragraph{Training} 
As our method is capable of adopting initial parameters from a pre-trained model, we re-use publicly available checkpoints of POMO (details in Appendix~\ref{sec:appendix-checkpoints}) as the starting point for \compass on TSP and CVRP. For JSSP, we found attention-based model from~\cite{jumanji2023github} outperforms L2D and hence choose it to be the reference single-agent architecture. We train the model and use the same trained checkpoints for all methods. We then train \compass until convergence on the same training distribution as that used to train the initial checkpoint.  For TSP and CVRP these are problem instances with 100 locations uniformly sampled within a unit square.  For JSSP, we use the same training distribution used in EAS, which is an instance with 10 jobs and machines, and a maximum possible duration of 98.
A single set of hyperparameters is used across all problems, with full training details provided in Appendix~\ref{sec:appendix-hp}.

\paragraph{Inference}
When evaluating active-search performance, each method is given a fixed budget of 1600 attempts -- similar to~\citet{hottung2022efficient, poppy} --, where each attempt consists of one trajectory per possible starting point. This approach is used to enable direct comparison to POMO and EAS which use rollouts from all starting points at each step. For the main results on TSP and CVRP, we do not use the ``augmentation trick''; where the same problem is solved multiple times by rotating the coordinate frame to make it appear different and thus generate additional diverse trajectories. This trick was used in a few baselines from prior work, however, we refrain from using it in the main results of this work for two reasons: (1) it is a domain-specific trick mainly applicable to routing problems and (2) it significantly increases the required computational budget. We nevertheless provide some results in both settings to ease comparison with previous work. Overall, the trajectory budget is exactly the same as the one used in~\citet{poppy, hottung2022efficient}. Note that expressing the budget in terms of trajectories gives an advantage to EAS, which uses more time, memory, and computation due to the backpropagations used to update the policy during the search.

\paragraph{Code availability} We release the code\footnote{Code, checkpoints and evaluation sets are available at \url{https://github.com/instadeepai/compass}} used to train our method and to run all baselines. We also make our checkpoints available for all three problems, along with the datasets necessary to reproduce the results. To ensure fair comparison and extend our evaluation to new settings, we reimplemented all baselines within the same codebase. For the three problems, we used the JAX~\citep{jax2018} implementations from Jumanji~\citep{jumanji2023github} to leverage hardware accelerators (e.g. TPU). Our code is optimized for TPU v3-8, which is the hardware used for our experiments.

\subsection{Standard benchmarking on TSP, CVRP, and JSSP}
\label{sec:larger_instances}

We evaluate our method on benchmark sets frequently used in the literature~\citep{Kool2019, POMO, poppy, hottung2022efficient}. Specifically, for TSP and CVRP, we use datasets of \num{10000} instances drawn from the training distribution, with the positions of \num{100} cities/customers uniformly sampled within the unit square, and three datasets not seen during training, each containing \num{1000} problem instances but with larger sizes: \num{125}, \num{150} and \num{200}, also generated from a uniform distribution over the unit square. We use the exact same datasets as in the literature.

\begin{table}[t]
  \centering
\caption{Results of \compass against the baseline algorithms for (a) TSP, (b) CVRP, and (c) JSSP problems. The methods are evaluated on instances from training distribution as well as on larger instance sizes to test generalization.}
  \label{fig:benchm_results}
\begin{subtable}[b]{\textwidth}
\caption{TSP}
  \centering
    \scalebox{0.71}{
    \begin{tabular}{l | ccc | ccc | ccc | ccc |}
        & \multicolumn{3}{c|}{\textbf{Training distr.}}
        & \multicolumn{9}{c|}{\textbf{Generalization}} \\
      & \multicolumn{3}{c|}{$n=100$} & \multicolumn{3}{c|}{$n=125$} & \multicolumn{3}{c|}{$n=150$} & \multicolumn{3}{c|}{$n=200$} \\
    Method & Obj. & Gap & Time & Obj. & Gap & Time & Obj. & Gap & Time & Obj. & Gap & Time \\
    \midrule
    Concorde & 7.765 & $0.000\%$ & 82M & 8.583 & $0.000\%$ & 12M & 9.346 & $0.000\%$ & 17M & 10.687 & $0.000\%$ & 31M \\
    LKH3 & 7.765 & $0.000\%$ & 8H & 8.583 & $0.000\%$ & 73M & 9.346 & $0.000\%$ & 99M & 10.687 & $0.000\%$ & 3H \\
    \midrule
    
    \begin{tabular}{@{}ll@{}}
    POMO (greedy)\\
    POMO (sampling)\\
    \poppy 16 \\
    EAS \\
    \textbf{\compass (ours)} \\
    \end{tabular} &

    \begin{tabular}{@{}c@{}}
    7.796\\
    7.779\\
    7.766\\
    7.778\\
    \textbf{7.765} \\
    \end{tabular} &
    
    \begin{tabular}{@{}c@{}}
    0.404\% \\
    0.185\% \\
    0.013\% \\
    0.161\% \\
    \textbf{0.002\%} \\
    \end{tabular} &

    \begin{tabular}{@{}c@{}}
    37S \\
    2H \\
    2H \\
    3H \\
    2H \\
    \end{tabular} &

    \begin{tabular}{@{}c@{}}
    8.635 \\
    8.609 \\
    8.587\\
    8.604\\
    \textbf{8.586} \\
    \end{tabular} &
    
    \begin{tabular}{@{}c@{}}
    0.607\% \\
    0.299\% \\
    0.050\% \\
    0.238\% \\
    \textbf{0.036\%} \\
    \end{tabular} &

    \begin{tabular}{@{}c@{}}
    6S \\
    20M \\
    20M \\
    38M \\
    20M \\
    \end{tabular} &

    \begin{tabular}{@{}c@{}}
    9.440 \\
    9.401 \\
    9.359\\
    9.380\\
    \textbf{9.354} \\
    \end{tabular} &
    
    \begin{tabular}{@{}c@{}}
    1.001\% \\
    0.585\% \\
    0.141\%\\
    0.363\%\\
    \textbf{0.083\%} \\
    \end{tabular} &

    \begin{tabular}{@{}c@{}}
    10S \\
    32M \\
    32M \\
    1H \\
    32M \\
    \end{tabular} &

    \begin{tabular}{@{}c@{}}
    10.933 \\
    10.956 \\
    10.795\\
    10.759\\
    \textbf{10.724} \\
    \end{tabular} &
    
    \begin{tabular}{@{}c@{}}
    2.300\% \\
    2.513\% \\
    1.007\%\\
    0.672\%\\
    \textbf{0.348\%} \\
    \end{tabular} &

    \begin{tabular}{@{}c@{}}
    21S \\
    70M \\
    70M \\
    101M \\
    70M \\
    \end{tabular}

    \end{tabular}}
    \label{tab:tsp_size}
\end{subtable}
\hfill
\begin{subtable}[b]{\textwidth}
\caption{CVRP}
    \centering
    \scalebox{0.68}{
    \begin{tabular}{l | ccc | ccc | ccc | ccc |}
        & \multicolumn{3}{c|}{\textbf{Training distr.}}
        & \multicolumn{9}{c|}{\textbf{Generalization}} \\
      & \multicolumn{3}{c|}{$n=100$} & \multicolumn{3}{c|}{$n=125$} & \multicolumn{3}{c|}{$n=150$} & \multicolumn{3}{c|}{$n=200$} \\
    Method & Obj. & Gap & Time & Obj. & Gap & Time & Obj. & Gap & Time & Obj. & Gap & Time \\
    \midrule
    LKH3 & 15.65 & $0.000\%$ & - & 17.50 & $0.000\%$ & - & 19.22 & $0.000\%$ & - & 22.00 & $0.000\%$ & - \\
    \midrule
    
    \begin{tabular}{@{}ll@{}}
    POMO (greedy)\\
    POMO (sampling)\\
    \poppy 32 \\
    EAS \\
    \textbf{\compass (ours)} \\
    \end{tabular} &

    \begin{tabular}{@{}c@{}}
    15.874\\
    15.713\\
    15.663\\
    15.663\\
    \textbf{15.594} \\
    \end{tabular} &
    
    \begin{tabular}{@{}c@{}}
    1.430\% \\
    0.399\% \\
    0.084\% \\
    0.081\% \\
    \textbf{-0.361\%} \\
    \end{tabular} &

    \begin{tabular}{@{}c@{}}
    2M \\
    4H \\
    4H \\
    7H \\
    4H \\
    \end{tabular} &

    \begin{tabular}{@{}c@{}}
    17.818 \\
    17.612 \\
    17.548\\
    17.536\\
    \textbf{17.511}\\
    \end{tabular} &
    
    \begin{tabular}{@{}c@{}}
    1.818\% \\
    0.642\% \\
    0.276\% \\
    0.146\% \\
    \textbf{0.064\%} \\
    \end{tabular} &

    \begin{tabular}{@{}c@{}}
    <1M \\
    43M \\
    42M \\
    81M \\
    42M \\
    \end{tabular} &

    \begin{tabular}{@{}c@{}}
    19.750 \\
    19.488 \\
    19.421 \\
    19.321 \\
    \textbf{19.313} \\
    \end{tabular} &

    \begin{tabular}{@{}c@{}}
    2.757\% \\
    1.393\% \\
    1.044\% \\
    0.528\% \\
    \textbf{0.485\%} \\
    \end{tabular} &

    \begin{tabular}{@{}c@{}}
    1M \\
    1H \\
    1H \\
    2H \\
    1H \\
    \end{tabular} &

    \begin{tabular}{@{}c@{}}
    23.318\\
    23.378 \\
    23.352\\
    22.541 \\
    \textbf{22.462}\\
    \end{tabular} &
    
    \begin{tabular}{@{}c@{}}
    5.992\% \\
    6.264\% \\
    6.144\% \\
    2.460\% \\
    \textbf{2.098\%} \\
    \end{tabular} &

    \begin{tabular}{@{}c@{}}
    2M \\
    100M \\
    100M \\
    3H \\
    100M \\
    \end{tabular}

    \end{tabular}}
    \label{tab:cvrp_size}
\end{subtable}
\hfill
\begin{subtable}[b]{0.9\textwidth}
\caption{JSSP}
    \centering
    \scalebox{0.8}{
      \begin{tabular}{l | ccc | ccc | ccc |}
        & \multicolumn{3}{c|}{\textbf{Training distr.}}
        & \multicolumn{6}{c|}{\textbf{Generalization}} \\
      & \multicolumn{3}{c|}{$10 \times 10$} & \multicolumn{3}{c|}{$15 \times 15$} & \multicolumn{3}{c|}{$20 \times 15$} \\
    Method & Obj. & Gap & Time & Obj. & Gap & Time & Obj. & Gap & Time \\
    \midrule
    OR-Tools & 807.6 & $0.0\%$ & 37S & 1188.0 & $0.0\%$ & 3H & 1345.5 & $0.0\%$ & 80H \\
    \midrule

    \begin{tabular}{@{}ll@{}}
    L2D (sampling) \\
    Single \\
    Poppy 16 \\
    EAS \\
    \textbf{\compass (ours)} \\
    \end{tabular} &
        \begin{tabular}{@{}c@{}}
    871.7 \\
    862.1 \\
    849.7 \\
    858.4 \\
    \textbf{845.5} \\
    \end{tabular} & 
        \begin{tabular}{@{}c@{}}
    $8.0\%$\\
    $6.7\%$\\
    $5.2\%$\\
    $6.3\%$\\
    \textbf{4.7\%}\\
    \end{tabular} & 

    \begin{tabular}{@{}c@{}}
    8H\\
    3H\\
    3H\\
    5H\\
    3H\\
    \end{tabular} &
    
    \begin{tabular}{@{}c@{}}
    1378.3\\
    1302.6 \\
    1290.4\\
    1295.2\\
    \textbf{1282.8}\\
    \end{tabular} & 
    
    \begin{tabular}{@{}c@{}}
    $16.0\%$\\
    $9.6\%$\\
    $8.6\%$\\
    $9.0\%$\\
    \textbf{8.0\%}\\
    \end{tabular} & 

    \begin{tabular}{@{}c@{}}
    25H\\
    5H\\
    5H\\
    9H\\
    5H\\
    \end{tabular} &
    
    \begin{tabular}{@{}c@{}}
    1624.6\\
    1503.0\\
    1495.7\\
    1498.0\\
    \textbf{1485.6}\\
    \end{tabular} & 
    
    \begin{tabular}{@{}c@{}}
    $20.8\%$\\
    $11.7\%$\\
    $11.2\%$\\
    $11.3\%$\\
    \textbf{10.4\%}\\
    \end{tabular} &

    \begin{tabular}{@{}c@{}}
    40H\\
    8H\\
    8H\\
    11H\\
    8H\\
    \end{tabular}
    
    \\

    \end{tabular}}
    \label{tab:jssp_size}
\end{subtable}
\end{table}

\paragraph{Results} The average performance of each method across all problem settings is presented in~\cref{fig:benchm_results}. We find that \compass demonstrates superior performance on all of the 11 test sets considered. Moreover, the degree of improvement is significant across all problem types. On TSP and JSSP, \compass reduces the optimality gap on the training distribution by a factor of 6.5 and 1.3, respectively. On CVRP, \compass is the only RL method able to outperform the industrial solver LKH. Finally, \compass is also found to generalize well to larger problem instances unseen during training. \compass obtains the best solutions in all TSP, CVRP and JSSP sets.

The same benchmark is also reported with the ``augmentation trick'' in~\cref{fig:benchm_results_augm} for TSP and CVRP. This trick can only be used for routing problems and is not applicable to JSSP. Interestingly, \compass is the only method that performs on par or better without the ``augmentation trick'', showing its ability to adapt and find diversity in its latent space rather than through a problem-specific augmentation.
In this setting, \compass is outperformed by EAS on two instance sizes of CVRP. Nevertheless, EAS is 50\% slower and more computationally expensive as it requires updating an entire subset of its network's weights, as opposed to simply navigating the 16-dimensional latent space of policies as is done in \compass. Overall, \compass remains the leading method and the conclusions drawn above remain unchanged.

\subsection{Robustness to generalization: solving mutated instances}
\label{sec:mutated_sets}

To further study the generalization ability of our method, we consider the mutation operators introduced by~\citet{diverse_tsp_instances} to procedurally transform instances drawn from the training distribution.
By progressively increasing the power of the applied mutations we construct new datasets that are increasingly far from the training distribution whilst not modifying the overall size of the problem.

We use 9 different mutation operators (explosion, implosion, cluster, rotation, linear projection, axis projection, expansion, compression and grid). One can find an illustration of the entire set of mutations along with their mathematical definition in Appendix~\ref{sec:appendix-mutations}. Interestingly, it enables us to evaluate the methods on instances that look closer to real-life situations. For instance, the operator that gathers nodes in a cluster can mimic a dense city surrounded by its nearby suburbs. In practice, each mutation operator is parameterized by a factor that controls the probability of mutating each node of the instance - referred to as mutation power - this factor directly impacts the shift between the training distribution and the new distribution. We use 10 values, going from 0 (no change) to 0.9 (highly mutated instances).

\begin{figure}
    \centering
    \includegraphics[width=0.7\linewidth]{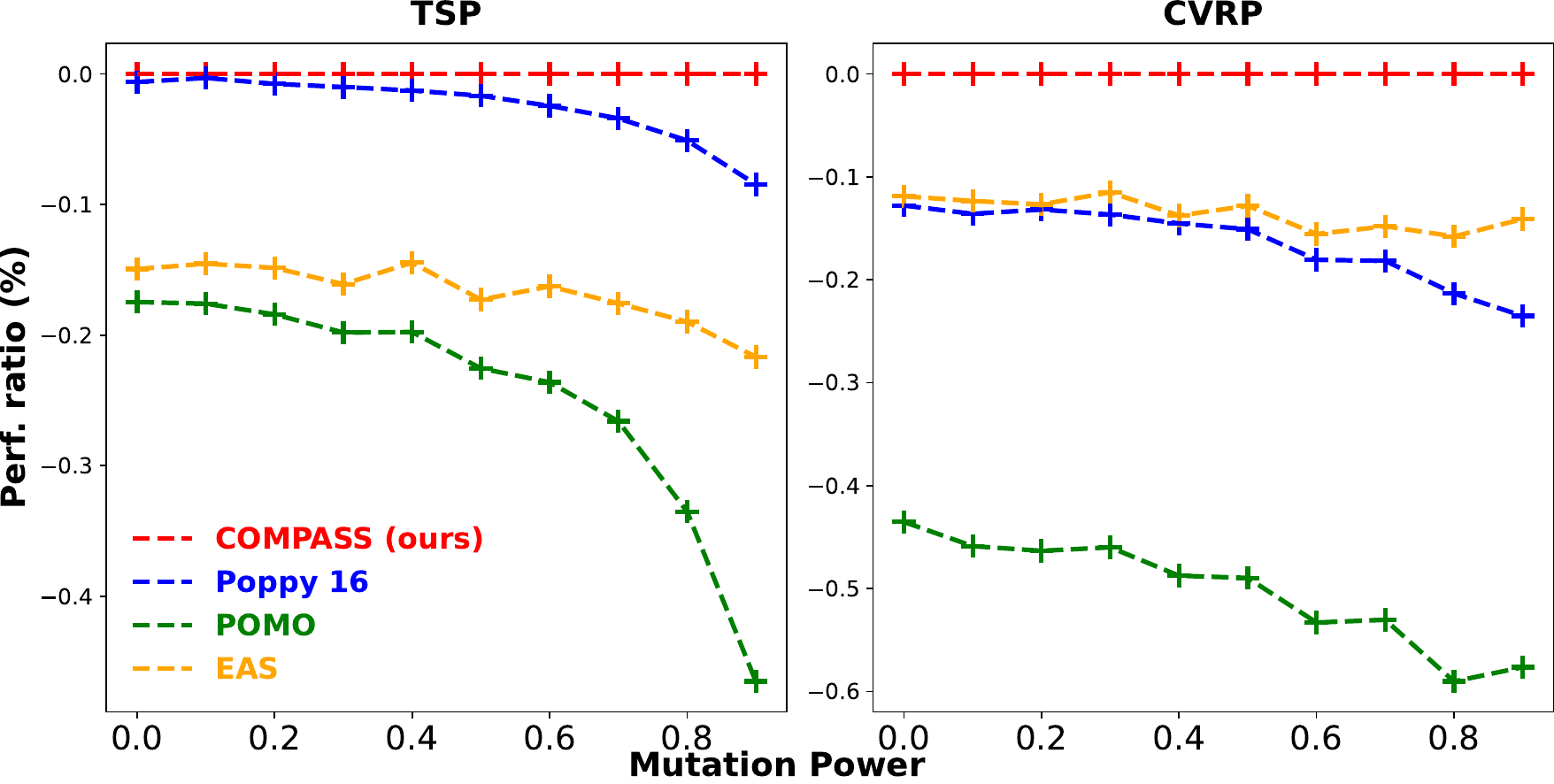}
    \caption{Relative difference between \compass and baselines as a function of mutation power. \compass outperforms the baselines on all 18 evaluation sets. Most methods have a decreasing performance ratio, showing that \compass generalizes better: its evolution strategy is able to find areas of its latent space that are high-performing, even on instances that are out-of-distribution.}
    \label{fig:distrib-shift-generalisation}
\end{figure}

\paragraph{Results}
We plot the relative performance of the baselines compared to \compass in \cref{fig:distrib-shift-generalisation}. A negative performance ratio indicates that a method does not provide as good of a solution as \compass, and we observe that this is the case for all baseline methods, at all mutation strengths, on both TSP and CVRP. Moreover, \compass is seen to generalize significantly better than the methods that only rely on stochastic sampling for their search, namely POMO and Poppy. This validates our intuition that adaptive policies are especially important for handling out-of-distribution data, where the optimal policy may be significantly different to that needed during pre-training. Even compared to EAS, which fine-tunes the policy to the target problem instance, we find that \compass maintains a significant performance gap across all mutation strengths. This result is particularly noteworthy as our approach only modifies 16 parameters (the conditioning vector sampled from our latent space), compared to EAS, which updates more than $10^4$ parameters (the embeddings of the instance's nodes).

It is interesting to note that the relative generalization performance of \compass compared to EAS is stronger on these mutated instances than the larger instances considered in \cref{sec:larger_instances}.  We hypothesize that this is because EAS actively fine-tunes the embeddings of every location in a given problem instance. Therefore, as the problem size increases, so does the number of free parameters to adapt the policy (albeit with commensurately increasing computational overhead).  This suggests that further improvements to \compass could be possible by increasing the number of adapted parameters (i.e.\ the latent space dimension), however, we defer further investigation to future works.

\subsection{Analysis of the search strategies}
\label{sec:exp-search-strategies}
\begin{figure}
    \centering
    \includegraphics[width=.7\linewidth]{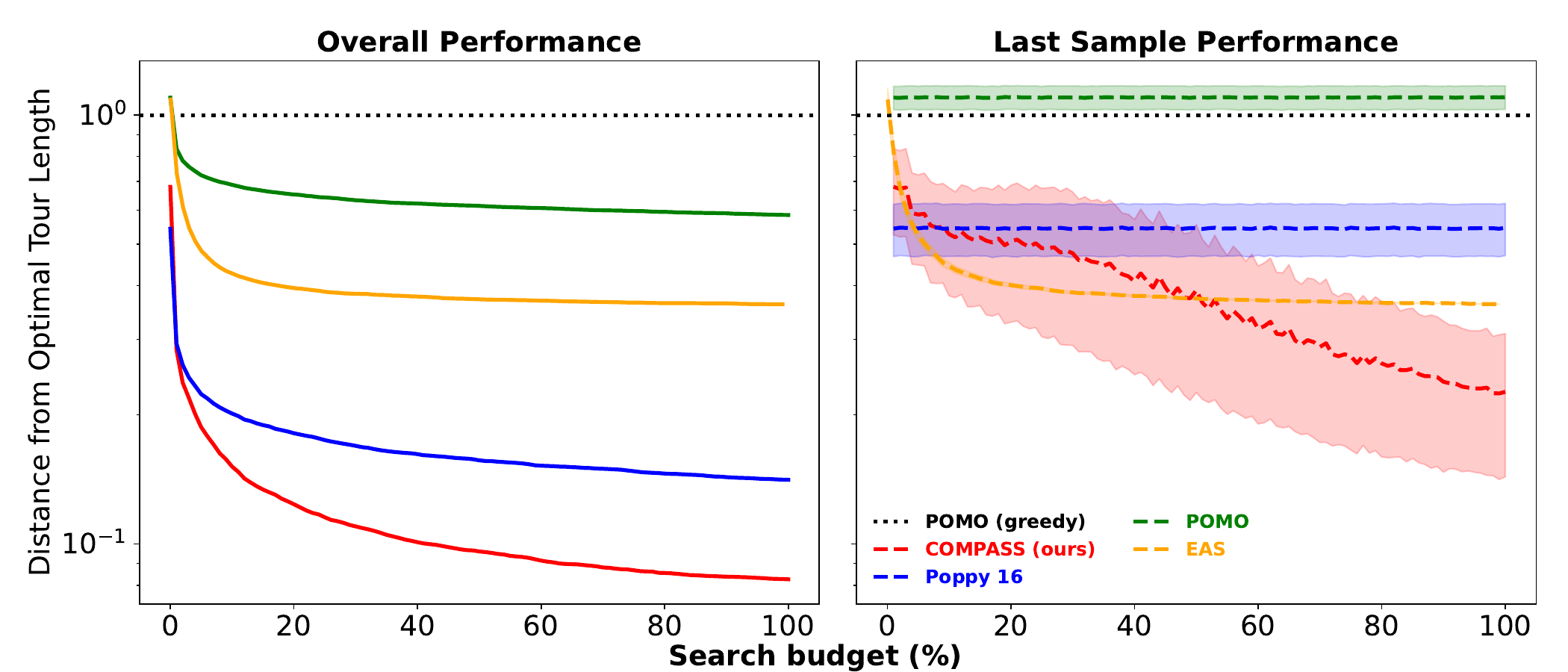}
    \caption{Evolution of the overall performance and last performance obtained by the methods during their search on TSP150 - averaged on 1000 instances. The right plot reports mean and standard deviations of the most recent shots tried by methods during the search. It illustrates how \compass efficiently explores its latent space to search for high-performing solutions.}
    \label{fig:search-phase-plots}
\end{figure}

\begin{wrapfigure}{r}{0.4\textwidth}
    \vspace{-1.5\baselineskip}
    \centering
    \includegraphics[width=\linewidth]{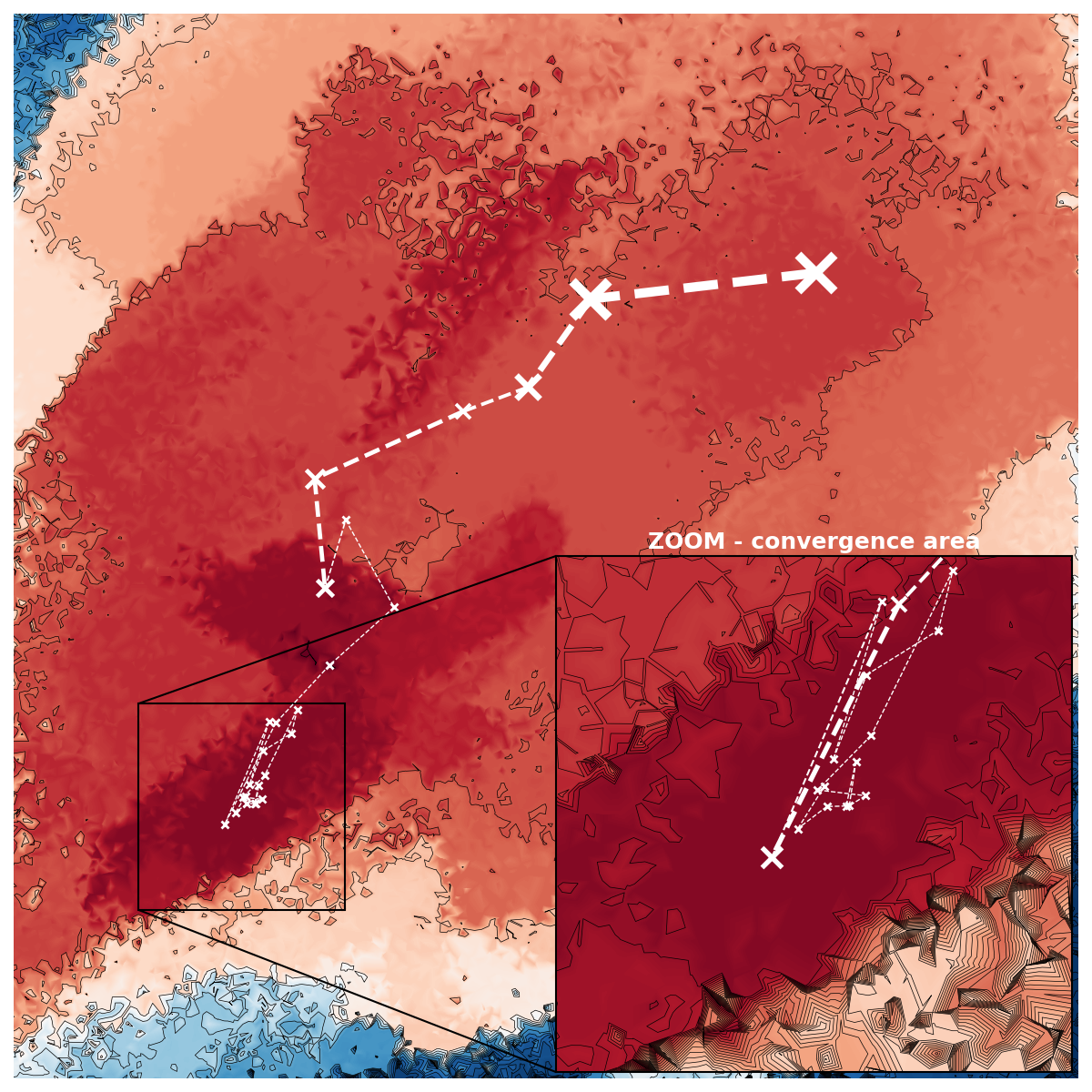}
    \caption{Contour plot of \compass's latent space, reflecting performance on a problem instance. White crosses show the successive means of a CMA-ES component during the search. The width of the path is proportional to the search's variance.
    }
    \label{fig:cmaes-search}
    \vspace{-1\baselineskip}
\end{wrapfigure}

In this section, we analyze the structure of the latent space and the behavior of the search procedure both empirically and visually.

\Cref{fig:search-phase-plots} details the performance of our considered methods as a function of the overall search budget. 
The left panel reports the quality of the best solution found so far (i.e.\ the cumulative performance), whereas the right plot reports the mean and standard deviations of the latest batch of solutions (i.e.\ the current performance) during the search process.
From this, we would highlight three main conclusions.
(\romannumeral 1)~Adaptive methods (\compass, EAS) perform well as they are able to improve the mean performance of their solution over time, in contrast to stochastic sampling methods (Poppy, POMO). This also highlights that the latent space of \compass has been able to diversify and can be exploited.
(\romannumeral 2)~Highly-focused (low-variance) search does not always outperform stochastic exploration.  Concretely, whilst EAS quickly adapts a policy with better average performance than Poppy (right panel), the additional variance of Poppy's multiple diverse policies means it produces better overall solutions (left panel).
(\romannumeral 3)~\compass is able to combine both of the previously discussed aspects for a highly performant search procedure.  By using an adaptive covariance mechanism as well as its multiple components to navigate several regions of the latent policy space, it focuses its search on promising strategies (better average performance) whilst maintaining a broad beam (higher variance).

To better understand how \compass's latent space is structured and explored, \cref{fig:cmaes-search} presents the trajectory of a single CMA-ES component during the search of a 2D latent space on a randomly chosen problem instance. We can first observe that even for a specific problem instance, there are several high-performing areas of interest which highlights the advantage of having multiple search components. Furthermore, it shows how the evolution strategy explores the space. The search variance is initially high to improve exploration until the search center moves into a high-performing area, whereupon the variance is gradually decreased to better exploit these promising strategies. We provide additional plots and explanation in~\cref{sec:appendix-latent-space-plots} for other problem instances, demonstrating the spread of the specialised areas depending on the problem instance.

Lastly, it is worth noting that the adaptation mechanism of COMPASS (CMA-ES search) comes with negligible time cost (e.g. three orders of magnitude smaller than the time needed for the environment rollout), which is a strength compared to the costly backpropagation-based updates performed in EAS. We provide additional time analysis in~\cref{appendix:time-consumption}.

\begin{table}[h!]
    \centering
    \caption{Results of COMPASS and the baseline algorithms with instance augmentation for (a) TSP and (b) CVRP. We also report COMPASS with no augmentation (no aug.).}
    \label{fig:benchm_results_augm}
    \begin{subtable}[b]{\textwidth}
    \caption{TSP}
      \centering
        \scalebox{0.58}{
        \begin{tabular}{l | ccc | ccc | ccc | ccc |}
            & \multicolumn{3}{c|}{\textbf{Training distr.}}
            & \multicolumn{9}{c|}{\textbf{Generalization}} \\
          & \multicolumn{3}{c|}{$n=100$} & \multicolumn{3}{c|}{$n=125$} & \multicolumn{3}{c|}{$n=150$} & \multicolumn{3}{c|}{$n=200$} \\
        Method & Obj. & Gap & {Time} & Obj. & Gap & {Time} & Obj. & Gap & {Time} & Obj. & Gap & {Time} \\
        \midrule
        
        \begin{tabular}{@{}ll@{}}
        CVAE-Opt \\
        SGBS \\
        SGBS+EAS-Lay \\
        POMO (sampling)\\
        Poppy 16 \\
        EAS \\
        \textbf{COMPASS (aug.)} \\
        COMPASS (no aug.) \\
        \end{tabular} &
    
        \begin{tabular}{@{}c@{}}
        - \\
        7.769 \\
        7.769 \\
        7.767 \\
        \textbf{7.765}\\
        7.768 \\
        \textbf{7.765} \\
        \textbf{7.765} \\
        \end{tabular} &
        
        \begin{tabular}{@{}c@{}}
        0.343\% \\
        0.058\% \\
        0.058\% \\
        0.026\% \\
        \textbf{0.002\%} \\
        0.038\% \\
        \textbf{0.002\%} \\
        \textbf{0.002\%} \\
        \end{tabular} &
    
        \begin{tabular}{@{}c@{}}
        6D \\
        9M \\
        3H \\
        2H \\
        2H \\
        3H \\
        2H \\
        2H \\
        \end{tabular} &
    
        \begin{tabular}{@{}c@{}}
        8.646 \\
        - \\
        - \\
        8.594 \\
        \textbf{8.584}\\
        8.590 \\
        \textbf{8.584} \\
        8.586 \\
        \end{tabular} &
        
        \begin{tabular}{@{}c@{}}
        0.736\% \\
        - \\
        - \\
        0.128\% \\
        \textbf{0.009\%} \\
        0.080\% \\
        \textbf{0.009\%} \\
        0.036\% \\
        \end{tabular} &
    
        \begin{tabular}{@{}c@{}}
        21H \\
        - \\
        - \\
        20M \\
        20M \\
        31M \\
        20M \\
        20M \\
        \end{tabular} &
    
        \begin{tabular}{@{}c@{}}
        9.482 \\
        9.367 \\
        9.359 \\
        9.376 \\
        9.351\\
        9.361 \\
        \textbf{9.350} \\
        9.354 \\
        \end{tabular} &
        
        \begin{tabular}{@{}c@{}}
        1.45\% \\
        0.220\% \\
        0.174\% \\
        0.321\% \\
        0.141\% \\
        0.159\% \\
        \textbf{0.043\%} \\
        0.083\% \\
        \end{tabular} &
    
        \begin{tabular}{@{}c@{}}
        30H \\
        8M \\
        1H \\
        32M \\
        32M \\
        50M \\
        32M \\
        32M \\
        \end{tabular} &
    
        \begin{tabular}{@{}c@{}}
        - \\
        10.753 \\
        10.727 \\
        10.916 \\
        10.802\\
        10.730\\
        \textbf{10.723} \\
        10.724 \\
        \end{tabular} &
        
        \begin{tabular}{@{}c@{}}
        - \\
        0.619\% \\
        0.40\% \\
        2.14\% \\
        1.08\%\\
        0.403\%\\
        \textbf{0.337\%} \\
        0.348\% \\
        \end{tabular} &
    
        \begin{tabular}{@{}c@{}}
        - \\
        14M \\
        3H \\
        70M \\
        70M \\
        85M \\
        70M \\
        70M \\
        \end{tabular}
    \end{tabular}}
    \label{tab:tsp_size_augm}
    \end{subtable}
    \hfill
    \begin{subtable}[b]{\textwidth}
    \caption{CVRP}
        \centering
        \scalebox{0.58}{
        \begin{tabular}{l | ccc | ccc | ccc | ccc |}
            & \multicolumn{3}{c|}{\textbf{Training distr.}}
            & \multicolumn{9}{c|}{\textbf{Generalization}} \\
          & \multicolumn{3}{c|}{$n=100$} & \multicolumn{3}{c|}{$n=125$} & \multicolumn{3}{c|}{$n=150$} & \multicolumn{3}{c|}{$n=200$} \\
        Method & Obj. & Gap & {Time} & Obj. & Gap & {Time} & Obj. & Gap & {Time} & Obj. & Gap & {Time} \\
        \midrule
        
        \begin{tabular}{@{}ll@{}}
        CVAE-Opt \\
        SGBS \\
        SGBS+EAS-Lay \\
        POMO (sampling) \\
        Poppy 32 \\
        EAS \\
        COMPASS (aug.) \\
        \textbf{COMPASS (no aug.)} \\
        \end{tabular} &
    
        \begin{tabular}{@{}c@{}}
        - \\
        15.66 \\
        \textbf{15.594} \\
        15.67\\
        15.62\\
        15.623 \\
        15.65 \\
        \textbf{15.594} \\
        \end{tabular} &
        
        \begin{tabular}{@{}c@{}}
        1.36\% \\
        0.01\% \\
        \textbf{-0.36\%} \\
        0.18\% \\
        -0.14\% \\
        -0.175\% \\
        -0.00\% \\
        \textbf{-0.36\%} \\
        \end{tabular} &
    
        \begin{tabular}{@{}c@{}}
        11D \\
        10M \\
        6H \\
        4H \\
        4H \\
        7H \\
        4H \\
        4H \\
        \end{tabular} &
    
        \begin{tabular}{@{}c@{}}
        17.87 \\
        - \\
        - \\
        17.56 \\
        17.49\\
        \textbf{17.473} \\
        17.52 \\
        17.511\\
        \end{tabular} &
        
        \begin{tabular}{@{}c@{}}
        2.08\% \\
        - \\
        - \\
        0.33\% \\
        -0.10\% \\
        \textbf{-0.153\%} \\
        0.09\% \\
        0.06\% \\
        \end{tabular} &
    
        \begin{tabular}{@{}c@{}}
        36H \\
        - \\
        - \\
        43M \\
        42M \\
        67M \\
        42M \\
        42M \\
        \end{tabular} &
    
        \begin{tabular}{@{}c@{}}
        19.84 \\
        19.43 \\
        \textbf{19.168} \\
        19.43 \\
        19.32 \\
        19.261 \\
        19.33 \\
        19.313 \\
        \end{tabular} &
    
        \begin{tabular}{@{}c@{}}
        3.24 \% \\
        1.08\% \\
        \textbf{-0.27\%} \\
        1.08\% \\
        0.50\% \\
        0.213\% \\
        0.56\% \\
        0.49\% \\
        \end{tabular} &
    
        \begin{tabular}{@{}c@{}}
        46H \\
        4M \\
        2H \\
        1H \\
        1H \\
        2H \\
        1H \\
        1H \\
        \end{tabular} &
    
        \begin{tabular}{@{}c@{}}
        - \\
        22.57 \\
        \textbf{21.988} \\
        23.24 \\
        22.94 \\
        22.556 \\
        22.55 \\
        22.462 \\
        \end{tabular} &
        
        \begin{tabular}{@{}c@{}}
        - \\
        2.59\% \\
        \textbf{-0.05\%} \\
        5.64\% \\
        4.27\% \\
        2.527\% \\
        2.49\% \\
        2.10\% \\
        \end{tabular} &
    
        \begin{tabular}{@{}c@{}}
        - \\
        9M \\
        5H \\
        100M \\
        100M \\
        3H \\
        100M \\
        100M \\
        \end{tabular}
    
        \end{tabular}}
        \label{tab:cvrp_size_augm}
    \end{subtable}
\end{table}
\vspace{-0.25cm}

%% file: sections/5-conclusion.tex
\section{Conclusion}

We present \compass, a novel approach to solving CO problems using RL. Our approach is motivated by the observation that active search is a key component to finding high-quality solutions to NP-hard problems. Finding one-shot solutions that are near-optimal is believed to be impossible. Instead, \compass is trained to create a distribution of diverse and specialized policies, conditioned on a structured latent space. This diversification is achieved by using an objective that specializes areas of the space on sub-distributions of problem instances. By navigating this latent space at inference time \compass is able to find the most performant policy for a given instance.  Empirical results show that \compass achieves state-of-the-art performance on all 11 standard benchmark tasks across three distinct CO problems, TSP, CVRP and JSSP, outperforming prior RL methods based on either stochastic sampling or fine-tuning. We extend the canonical evaluation sets with instances that are procedurally transformed using mutation operators introduced in prior work. This additional set of tasks enables us to assess the generalization ability of \compass. We show that \compass is particularly robust to out-of-distribution instances, achieving superior performance in all 18 tasks considered. To better understand the benefits of our search procedure, we provide an empirical analysis of the latent space's structure, along with evidence of how it is explored at inference time. We show that, despite having no explicit regularization during training, the latent space exhibits clear regions of interest, and our search procedure is able to explore this space using an evolution strategy to produce high-performing policies. Overall, \compass proves to be performant, robust, and versatile on many types of CO problems and is able to provide solutions quickly at a reasonable computational cost.

\textbf{Limitations and future work.} The diversity of the policies contained in the latent space is closely linked to the specialisation that can be obtained from the training distribution, and hence potentially limited. We would like to inspect whether a broader range of policies could be obtained by using an additional unsupervised diversity reward, or by procedurally diversifying the distribution used. Another limitation of our method is the lack of structure in our latent space. Although we proved that it was enough to be successfully explored by an evolution strategy, we hypothesize that a better defined space could be searched through faster. We would like to inspect the use of regularization terms during the training phase to achieve this.

%% file: sections/99-appendix.tex
\section{Extended Results}
\label{sec:appendix-all-tables}

In this section, we compare the results of our method to an extensive list of baselines for the TSP, CVRP, and JSSP problems. Subsection~\ref{subsec:larger-instances} focuses on the results of the standard benchmark, corresponding to instances from the training distribution as well as larger instances. Subsection~\ref{subsec:ood-instances} presents the results for procedurally transformed (and hence out-of-distribution) instances.

\subsection{Extended results on standard benchmark}
\label{subsec:larger-instances}

In Section~\ref{experiments} of the main paper, we report the performances of the main competitors on a standard benchmark for TSP, CVRP, and JSSP. We report extended results, with other competitors (2-Opt-DL~\citep{wu2021learning} and LIH~\citep{Costa2020}) and inference time. We also report the results of an alternative architecture (L2D) for JSSP. Tables \ref{table:TSP_results}, \ref{table:CVRP_results}, and \ref{table:JSSP_results} show the results of all methods for TSP, CVRP, and JSSP problems, respectively. The first column in the tables shows the average performance across the validation set, which is the  mean tour length for TSP and CVRP, and the mean schedule duration for JSSP, and the second and third columns show the optimality gap and total run-time, respectively.

Examining the inference time provides valuable insights for comparison, especially considering that time constraints are often a crucial factor in industrial applications. On the whole benchmark, our method is as fast as the baselines (POMO, Poppy) and almost three times faster than EAS.

Results from Table~\ref{tab:jssp_size_full} confirm our choice to use the attention-based model from Jumanji~\citep{jumanji2023github}: the attention-based model is performing better on all sets, while being 5 times faster. We can also compare EAS used with both architectures (L2D or the attention-based model). EAS is performing better with the attention-based architecture in 2 datasets out of 3. Interestingly, EAS (w/ L2D) is performing very well on the training distribution, but generalizes less than EAS (w/ attention-based model).

\begin{table}[t]
  \centering
\caption{Results of \compass against the baseline algorithms for (a) TSP, (b) CVRP, and (c) JSSP problems. The methods are evaluated on instances from training distribution as well as on larger instance sizes to test generalization. Tables report the best solutions found, gaps to best industrial solvers, and inference times.}
\label{tab:additional_results}
  
\begin{subtable}[b]{0.9\textwidth}
\caption{TSP}
  \centering
    \scalebox{0.65}{
    \begin{tabular}{l | ccc | ccc | ccc | ccc |}
        & \multicolumn{3}{c|}{\textbf{Training distr.}}
        & \multicolumn{9}{c|}{\textbf{Generalization}} \\
      & \multicolumn{3}{c|}{$n=100$} & \multicolumn{3}{c|}{$n=125$} & \multicolumn{3}{c|}{$n=150$} & \multicolumn{3}{c|}{$n=200$} \\
    Method & Obj. & Gap & Time & Obj. & Gap & Time & Obj. & Gap & Time & Obj. & Gap & Time \\
    \midrule
    Concorde & 7.765 & $0.000\%$ & 82M & 8.583 & $0.000\%$ & 12M & 9.346 & $0.000\%$ & 17M & 10.687 & $0.000\%$ & 31M \\
    LKH3 & 7.765 & $0.000\%$ & 8H & 8.583 & $0.000\%$ & 73M & 9.346 & $0.000\%$ & 99M & 10.687 & $0.000\%$ & 3H \\
    \midrule
    
    \begin{tabular}{@{}ll@{}}
    2-Opt-DL \\
    LIH \\
    POMO (greedy)\\
    POMO\\
    \poppy 16 \\
    EAS \\
    \textbf{\compass (ours)} \\
    \end{tabular} &

    \begin{tabular}{@{}c@{}}
    7.83 \\
    7.87 \\
    7.796\\
    7.779\\
    7.766\\
    7.778\\
    \textbf{7.765} \\
    \end{tabular} &
    
    \begin{tabular}{@{}c@{}}
    0.87\% \\
    1.42\% \\
    0.404\% \\
    0.185\% \\
    0.013\% \\
    0.161\% \\
    \textbf{0.002\%} \\
    \end{tabular} &

    \begin{tabular}{@{}c@{}}
    - \\
    - \\
    41M \\
    2H \\
    2H \\
    3H \\
    2H \\
    \end{tabular} &

    \begin{tabular}{@{}c@{}}
    - \\
    - \\
    8.635 \\
    8.609 \\
    8.587\\
    8.604\\
    \textbf{8.586} \\
    \end{tabular} &
    
    \begin{tabular}{@{}c@{}}
    - \\
    - \\
    0.607\% \\
    0.299\% \\
    0.050\% \\
    0.238\% \\
    \textbf{0.036\%} \\
    \end{tabular} &

    \begin{tabular}{@{}c@{}}
    - \\
    - \\
    6S \\
    20M \\
    20M \\
    38M \\
    20M \\
    \end{tabular} &

    \begin{tabular}{@{}c@{}}
    - \\
    - \\
    9.440 \\
    9.401 \\
    9.359\\
    9.380\\
    \textbf{9.354} \\
    \end{tabular} &
    
    \begin{tabular}{@{}c@{}}
    - \\
    - \\
    1.001\% \\
    0.585\% \\
    0.141\%\\
    0.363\%\\
    \textbf{0.083\%} \\
    \end{tabular} &

    \begin{tabular}{@{}c@{}}
    - \\
    - \\
    10S \\
    32M \\
    32M \\
    1H \\
    32M \\
    \end{tabular} &

    \begin{tabular}{@{}c@{}}
    - \\
    - \\
    10.933 \\
    10.956 \\
    10.795\\
    10.759\\
    \textbf{10.724} \\
    \end{tabular} &
    
    \begin{tabular}{@{}c@{}}
    - \\
    - \\
    2.300\% \\
    2.513\% \\
    1.007\%\\
    0.672\%\\
    \textbf{0.348\%} \\
    \end{tabular} &

    \begin{tabular}{@{}c@{}}
    - \\
    - \\
    21S \\
    70M \\
    70M \\
    101M \\
    70M \\
    \end{tabular}

    \end{tabular}}
    \label{table:TSP_results}
\end{subtable}
\hfill
\begin{subtable}[b]{0.9\textwidth}
\caption{CVRP}
    \centering
    \scalebox{0.65}{
    \begin{tabular}{l | ccc | ccc | ccc | ccc |}
        & \multicolumn{3}{c|}{\textbf{Training distr.}}
        & \multicolumn{9}{c|}{\textbf{Generalization}} \\
      & \multicolumn{3}{c|}{$n=100$} & \multicolumn{3}{c|}{$n=125$} & \multicolumn{3}{c|}{$n=150$} & \multicolumn{3}{c|}{$n=200$} \\
    Method & Obj. & Gap & Time & Obj. & Gap & Time & Obj. & Gap & Time & Obj. & Gap & Time \\
    \midrule
    LKH3 & 15.65 & $0.000\%$ & - & 17.50 & $0.000\%$ & - & 19.22 & $0.000\%$ & - & 22.00 & $0.000\%$ & - \\
    \midrule
    
    \begin{tabular}{@{}ll@{}}
    LIH \\
    POMO (greedy)\\
    POMO\\
    \poppy 32 \\
    EAS \\
    \textbf{\compass (ours)} \\
    \end{tabular} &

    \begin{tabular}{@{}c@{}}
    16.03 \\
    15.874\\
    15.713\\
    15.663\\
    15.663\\
    \textbf{15.594} \\
    \end{tabular} &
    
    \begin{tabular}{@{}c@{}}
    2.47\% \\
    1.430\% \\
    0.399\% \\
    0.084\% \\
    0.081\% \\
    \textbf{-0.361\%} \\
    \end{tabular} &

    \begin{tabular}{@{}c@{}}
    5H \\
    2M \\
    4H \\
    4H \\
    7H \\
    4H \\
    \end{tabular} &

    \begin{tabular}{@{}c@{}}
    - \\
    17.818 \\
    17.612 \\
    17.548\\
    17.536\\
    \textbf{17.511}\\
    \end{tabular} &
    
    \begin{tabular}{@{}c@{}}
    - \\
    1.818\% \\
    0.642\% \\
    0.276\% \\
    0.146\% \\
    \textbf{0.064\%} \\
    \end{tabular} &

    \begin{tabular}{@{}c@{}}
    - \\
    <1M \\
    43M \\
    42M \\
    81M \\
    42M \\
    \end{tabular} &

    \begin{tabular}{@{}c@{}}
    - \\
    19.750 \\
    19.488 \\
    19.421 \\
    19.321 \\
    \textbf{19.313} \\
    \end{tabular} &

    \begin{tabular}{@{}c@{}}
    - \\
    2.757\% \\
    1.393\% \\
    1.044\% \\
    0.528\% \\
    \textbf{0.485\%} \\
    \end{tabular} &

    \begin{tabular}{@{}c@{}}
    - \\
    1M \\
    1H \\
    1H \\
    2H \\
    1H \\
    \end{tabular} &

    \begin{tabular}{@{}c@{}}
    - \\
    23.318\\
    23.378 \\
    23.352\\
    22.541 \\
    \textbf{22.462} \\
    \end{tabular} &
    
    \begin{tabular}{@{}c@{}}
    - \\
    5.992\% \\
    6.264\% \\
    6.144\% \\
    2.460\% \\
    \textbf{2.098\%} \\
    \end{tabular} &

    \begin{tabular}{@{}c@{}}
    - \\
    2M \\
    100M \\
    100M \\
    3H \\
    100M \\
    \end{tabular}

    \end{tabular}}
    \label{table:CVRP_results}
\end{subtable}
\hfill
\begin{subtable}[b]{0.9\textwidth}
\caption{JSSP}
    \centering
    \scalebox{0.8}{
      \begin{tabular}{l | ccc | ccc | ccc |}
        & \multicolumn{3}{c|}{\textbf{Training distr.}}
        & \multicolumn{6}{c|}{\textbf{Generalization}} \\
      & \multicolumn{3}{c|}{$10 \times 10$} & \multicolumn{3}{c|}{$15 \times 15$} & \multicolumn{3}{c|}{$20 \times 15$} \\
    Method & Obj. & Gap & Time & Obj. & Gap & Time & Obj. & Gap & Time \\
    \midrule
    OR-Tools & 807.6 & $0.0\%$ & 37S & 1188.0 & $0.0\%$ & 3H & 1345.5 & $0.0\%$ & 80H \\
    \midrule

    \begin{tabular}{@{}ll@{}}
    L2D (Greedy) \\
    L2D (Sampling) \\
    EAS (w/ L2D) \\
    Single \\
    Poppy 16 \\
    EAS \\
    \textbf{\compass (ours)} \\
    \end{tabular} &

    \begin{tabular}{@{}c@{}}
    988.6\\
    871.7 \\
    \textbf{837.0} \\
    862.1 \\
    849.7 \\
    858.4 \\
    845.5 \\
    \end{tabular} & 
    
    \begin{tabular}{@{}c@{}}
    $22.3\%$ \\
    $8.0\%$ \\
    \textbf{3.7\%} \\
    $6.7\%$ \\
    $5.2\%$ \\
    $6.3\%$ \\
    $4.7\%$ \\
    \end{tabular} & 

    \begin{tabular}{@{}c@{}}
    20S\\
    8H\\
    7H\\
    3H\\
    3H\\
    5H\\
    3H\\
    \end{tabular} &

    \begin{tabular}{@{}c@{}}
    1528.3\\
    1378.3\\
    1326.4 \\
    1302.6 \\
    1290.4\\
    1295.2\\
    \textbf{1282.8}\\
    \end{tabular} & 
    
    \begin{tabular}{@{}c@{}}
    $28.6\%$\\
    $16.0\%$\\
    $11.7\%$\\
    $9.6\%$\\
    $8.6\%$\\
    $9.0\%$\\
    \textbf{8.0\%}\\
    \end{tabular} & 

    \begin{tabular}{@{}c@{}}
    44S\\
    25H\\
    22H\\
    5H\\
    5H\\
    9H\\
    5H\\
    
    \end{tabular} &

    \begin{tabular}{@{}c@{}}
    1738.0\\
    1624.6\\
    1570.8\\
    1503.0\\
    1495.7\\
    1498.0\\
    \textbf{1485.6}\\
    \end{tabular} & 
    
    \begin{tabular}{@{}c@{}}
    $29.2\%$\\
    $20.8\%$\\
    $16.8\%$\\
    $11.7\%$\\
    $11.2\%$\\
    $11.3\%$\\
    \textbf{10.4\%}\\
    \end{tabular} &

    \begin{tabular}{@{}c@{}}
    60S \\
    40H\\
    37H\\
    8H\\
    8H\\
    11H\\
    8H\\
    \end{tabular}

    \end{tabular}}
    \label{table:JSSP_results}
\end{subtable}
\end{table}

\subsection{Results on the procedurally transformed instances}
\label{subsec:ood-instances}

In Section~\ref{sec:mutated_sets}, we present the performance of the methods under study on procedurally transformed instances. In particular, Figure~\ref{fig:distrib-shift-generalisation} reports the relative performance of the baselines compared to our method \compass. In this section, we report the numerical results corresponding to this plot. Tables~\ref{tab:TSP_ood_results} show the results of all methods on TSP and \ref{tab:CVRP_ood_results} on CVRP. Each line corresponds to a mutation power used to procedurally transform instances. Mutations are used to create 1800 new instances. The columns report the average tour length and the optimality gap. Details about the way those datasets are created can be found in Appendix~\ref{sec:appendix-mutations}.

Our method outperforms the baselines on the entire benchmark. Nevertheless, the industrial solver LKH3 is still performing better than \compass on all mutated datasets, showing room for improvement of the RL-based approaches.

\begin{table}[t]
  \centering
\caption{Results of the methods on procedurally transformed instances, obtained by applying mutations with increasing mutation powers (referred to as `Mu'). Our method \compass outperforms other baselines on all mutated instances.}
  \label{fig:mutated_instances_results}
\begin{subtable}[b]{0.9\textwidth}
\caption{TSP}
  \centering
    \scalebox{0.8}{
    \begin{tabular}{c | cc | cc | cc | cc | cc |}

      & \multicolumn{2}{c|}{LKH3} & \multicolumn{2}{c|}{POMO} & \multicolumn{2}{c|}{\poppy 16} & \multicolumn{2}{c|}{EAS} & \multicolumn{2}{c|}{\textbf{\compass (ours)}} \\
    \midrule
    Mu & Obj. & Gap & Obj. & Gap & Obj. & Gap & Obj. & Gap & Obj. & Gap \\
    \midrule
    
    \begin{tabular}{@{}ll@{}}
    0 \\
    1\\
    2\\
    3 \\
    4 \\
    5 \\
    6 \\
    7 \\
    8 \\
    9 \\
    \end{tabular} &

    \begin{tabular}{@{}c@{}}
    7.768\\
    7.609\\
    7.562\\
    7.485\\
    7.386\\
    7.308\\
    7.182\\
    7.063\\
    6.910\\
    6.732\\
    \end{tabular} &
    
    \begin{tabular}{@{}c@{}}
    0.000\% \\
    0.000\% \\
    0.000\% \\
    0.000\% \\
    0.000\% \\
    0.000\% \\
    0.000\% \\
    0.000\% \\
    0.000\% \\
    0.000\% \\
    \end{tabular} &

    \begin{tabular}{@{}c@{}}
    7.782\\
    7.624\\
    7.577\\
    7.502\\
    7.402\\
    7.326\\
    7.201\\
    7.085\\
    6.938\\
    6.770\\
    \end{tabular} &
    
    \begin{tabular}{@{}c@{}}
    0.188\% \\
    0.195\% \\
    0.199\% \\
    0.217\% \\
    0.219\% \\
    0.251\% \\
    0.272\% \\
    0.311\% \\
    0.402\% \\
    0.557\% \\
    \end{tabular} &

    \begin{tabular}{@{}c@{}}
    7.769\\
    7.611\\
    7.563\\
    7.488\\
    7.389\\
    7.311\\
    7.186\\
    7.069\\
    6.918\\
    6.744\\
    \end{tabular} &
    
    \begin{tabular}{@{}c@{}}
    0.02\% \\
    0.022\% \\
    0.022\% \\
    0.029\% \\
    0.034\% \\
    0.042\% \\
    0.06\% \\
    0.078\% \\
    0.118\% \\
    0.176\% \\
    \end{tabular} &

    \begin{tabular}{@{}c@{}}
    7.78\\
    7.622\\
    7.574\\
    7.499\\
    7.398\\
    7.322\\
    7.196\\
    7.079\\
    6.928\\
    6.753\\
    \end{tabular} &
    
    \begin{tabular}{@{}c@{}}
    0.163\% \\
    0.164\% \\
    0.163\% \\
    0.18\% \\
    0.166\% \\
    0.199\% \\
    0.199\% \\
    0.22\% \\
    0.256\% \\
    0.308\% \\
    \end{tabular} &

    \begin{tabular}{@{}c@{}}
    \textbf{7.769}\\
    \textbf{7.611}\\
    \textbf{7.563}\\
    \textbf{7.487}\\
    \textbf{7.388}\\
    \textbf{7.310}\\
    \textbf{7.184}\\
    \textbf{7.066}\\
    \textbf{6.914}\\
    \textbf{6.738}\\
    \end{tabular} &
    
    \begin{tabular}{@{}c@{}}
    \textbf{0.013\%} \\
    \textbf{0.019\%} \\
    \textbf{0.014\%} \\
    \textbf{0.019\%} \\
    \textbf{0.021\%} \\
    \textbf{0.026\%} \\
    \textbf{0.036\%} \\
    \textbf{0.044\%} \\
    \textbf{0.067\%} \\
    \textbf{0.091\%} \\
    \end{tabular}

    \end{tabular}}
    \label{tab:TSP_ood_results}
\end{subtable}
\hfill
\begin{subtable}[b]{0.9\textwidth}
\caption{CVRP}
    \centering
    \scalebox{0.8}{
    \begin{tabular}{c | cc | cc | cc | cc | cc |}

      & \multicolumn{2}{c|}{LKH3} & \multicolumn{2}{c|}{POMO} & \multicolumn{2}{c|}{\poppy 16} & \multicolumn{2}{c|}{EAS} & \multicolumn{2}{c|}{\textbf{\compass (ours)}} \\
    \midrule
    Mu & Obj. & Gap & Obj. & Gap & Obj. & Gap & Obj. & Gap & Obj. & Gap \\
    \midrule
    
    \begin{tabular}{@{}ll@{}}
    0 \\
    1 \\
    2 \\
    3 \\
    4 \\
    5 \\
    6 \\
    7 \\
    8 \\
    9 \\
    \end{tabular} &

    \begin{tabular}{@{}c@{}}
    15.595\\
    15.337\\
    15.271\\
    15.156\\
    15.032\\
    14.805\\
    14.561\\
    14.306\\
    13.886\\
    13.507\\
    \end{tabular} &
    
    \begin{tabular}{@{}c@{}}
    0.000\% \\
    0.000\% \\
    0.000\% \\
    0.000\% \\
    0.000\% \\
    0.000\% \\
    0.000\% \\
    0.000\% \\
    0.000\% \\
    0.000\% \\
    \end{tabular} &

    \begin{tabular}{@{}c@{}}
    15.661\\
    15.41\\
    15.345\\
    15.23\\
    15.112\\
    14.885\\
    14.646\\
    14.398\\
    13.989\\
    13.612\\
    \end{tabular} &
    
    \begin{tabular}{@{}c@{}}
    0.426\% \\
    0.471\% \\
    0.482\% \\
    0.488\% \\
    0.531\% \\
    0.545\% \\
    0.581\% \\
    0.641\% \\
    0.741\% \\
    0.773\% \\
    \end{tabular} &

    \begin{tabular}{@{}c@{}}
    15.613\\
    15.36\\
    15.294\\
    15.181\\
    15.061\\
    14.835\\
    14.594\\
    14.348\\
    13.936\\
    13.566\\
    \end{tabular} &
    
    \begin{tabular}{@{}c@{}}
    0.119\% \\
    0.148\% \\
    0.15\% \\
    0.164\% \\
    0.188\% \\
    0.206\% \\
    0.229\% \\
    0.292\% \\
    0.363\% \\
    0.431\% \\
    \end{tabular} &

    \begin{tabular}{@{}c@{}}
    15.611\\
    15.356\\
    15.29\\
    15.174\\
    15.058\\
    14.828\\
    14.587\\
    14.339\\
    13.922\\
    13.548\\
    \end{tabular} &
    
    \begin{tabular}{@{}c@{}}
    0.103\% \\
    0.121\% \\
    0.121\% \\
    0.123\% \\
    0.169\% \\
    0.154\% \\
    0.182\% \\
    0.227\% \\
    0.263\% \\
    0.298\% \\
    \end{tabular} &

    \begin{tabular}{@{}c@{}}
    \textbf{15.594}\\
    \textbf{15.339}\\
    \textbf{15.274}\\
    \textbf{15.16}\\
    \textbf{15.039}\\
    \textbf{14.813}\\
    \textbf{14.568}\\
    \textbf{14.322}\\
    \textbf{13.907}\\
    \textbf{13.534}\\
    \end{tabular} &
    
    \begin{tabular}{@{}c@{}}
    \textbf{-0.009\%} \\
    \textbf{0.012\%} \\
    \textbf{0.018\%} \\
    \textbf{0.027\%} \\
    \textbf{0.043\%} \\
    \textbf{0.055\%} \\
    \textbf{0.048\%} \\
    \textbf{0.11\%} \\
    \textbf{0.15\%} \\
    \textbf{0.196\%} \\
    \end{tabular}

    \end{tabular}}
    \label{tab:CVRP_ood_results}
\end{subtable}

\end{table}

\section{Analysis of the Performance during the Search Process}
\begin{figure}[h!]
  \centering
  \includegraphics[width=1.0\textwidth]{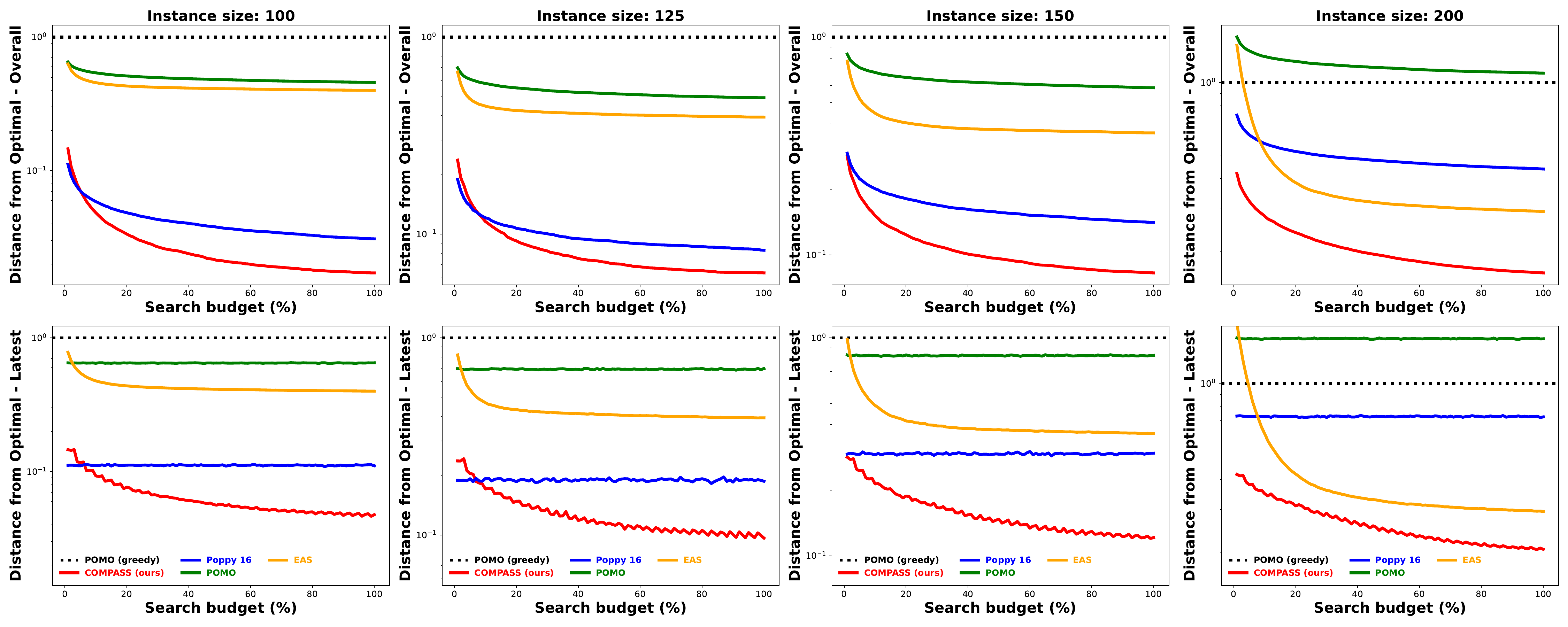}
  \caption{Evolution of the performance along the budget on \textbf{TSP}. This plots shows - for each method and each instance size - the evolution of (i) the performance of the method overall, hence the best solution ever found, on the top panel, and (ii) the performance of the latest sampled batch for each method, hence the best solution found in the recent attempts.}
  \label{fig:tsp-slowrl}
\end{figure}

\begin{figure}[h!]
  \centering
  \includegraphics[width=1.0\textwidth]{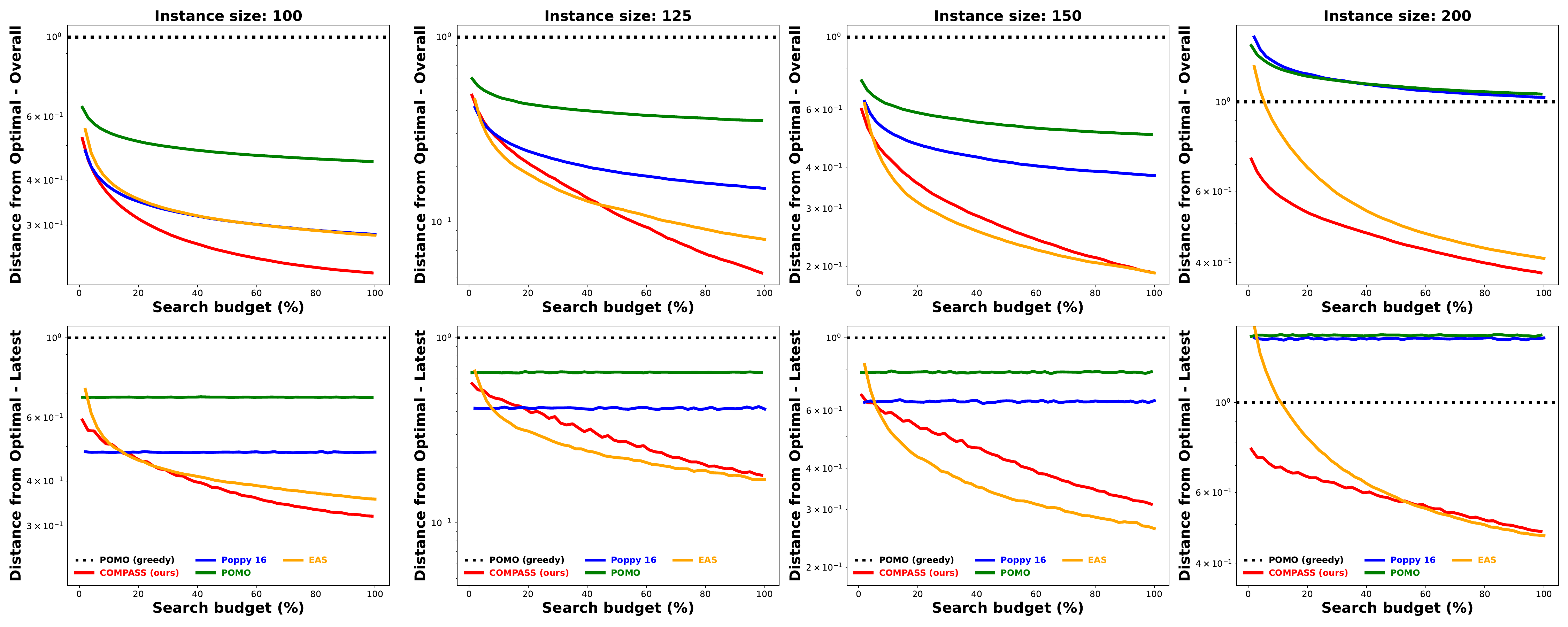}
  \caption{Evolution of the performance along the budget on \textbf{CVRP}. This plots shows - for each method and each instance size - the evolution of (i) the performance of the method overall, hence the best solution ever found, on the top panel, and (ii) the performance of the latest sampled batch for each method, hence the best solution found in the recent attempts.}
  \label{fig:cvrp-slowrl}
\end{figure}

\begin{figure}[h!]
  \centering
  \includegraphics[width=0.98\textwidth]{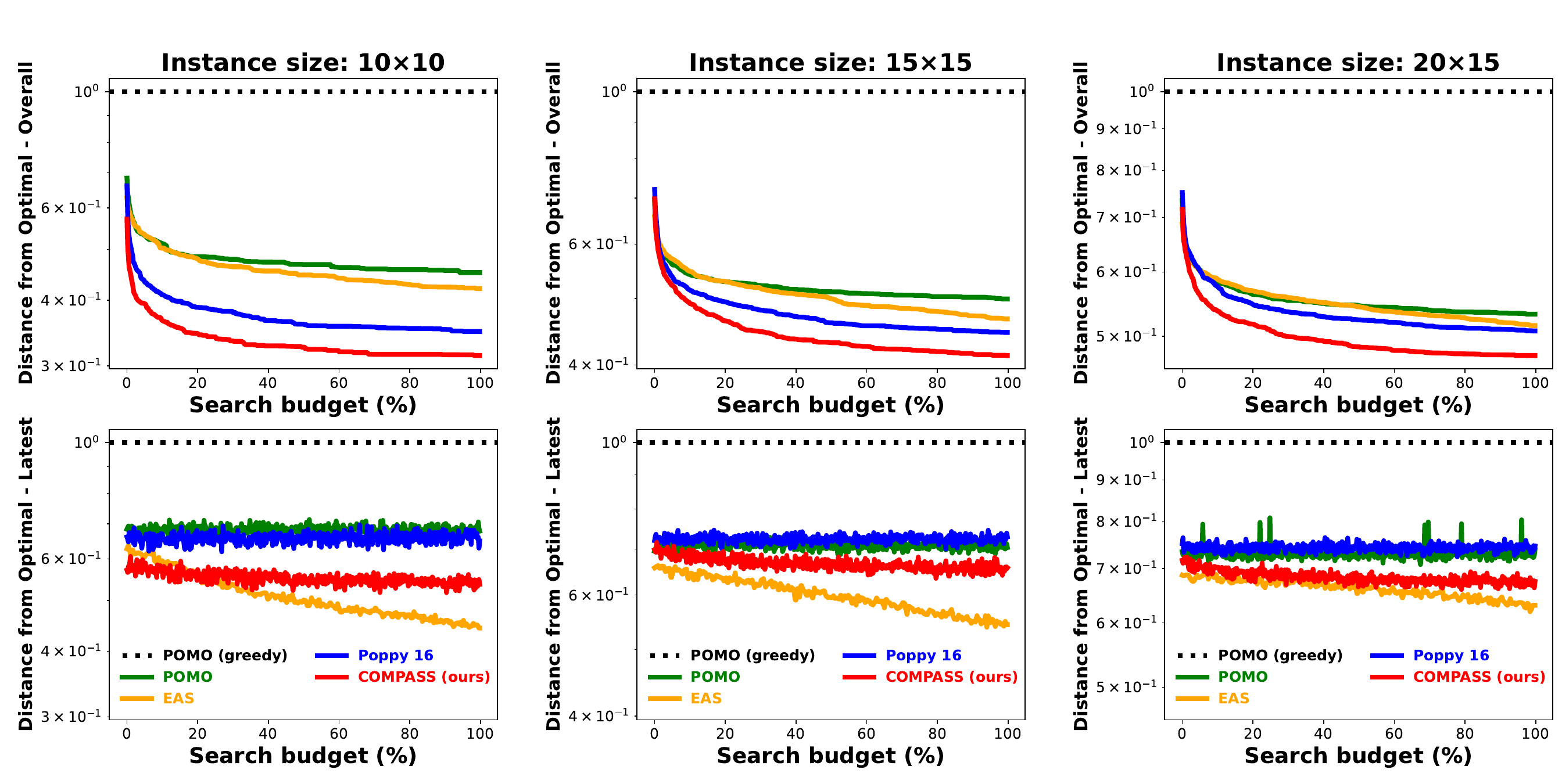}
  \caption{Evolution of the performance along the budget on \textbf{JSSP}. This plots shows - for each method and each instance size - the evolution of (i) the performance of the method overall, hence the best solution ever found, on the top panel, and (ii) the performance of the latest sampled batch for each method, hence the best solution found in the recent attempts.}
  \label{fig:jssp-slowrl}
\end{figure}

In this section, we examine how the performance of \compass, along with the baseline methods, evolves during the search process. The evaluation procedure consists of 160,000 rollouts per problem for TSP and CVRP, and 8,000 rollouts per problem for JSSP, distributed over the population for all problems. Figures~\ref{fig:tsp-slowrl}, \ref{fig:cvrp-slowrl}, \ref{fig:jssp-slowrl} show the performances -- for TSP, CVRP, and JSSP, respectively, -- for our method \compass and the three main baselines, POMO (single-agent for JSSP), Poppy, and EAS. Each figure showcases the overall performance and the latest performance achieved by the methods for various instance sizes, including the training distribution size (left column), medium size (middle column), and large size (right column). The first row of plots illustrates the performance of the best solution discovered thus far in the search process, while the second row of plots presents the best performance of the last batch of solutions found at the current timestep of the search process. 

Note that the `Latest' performance metric reported in these plots is different from the one reported in Section~\ref{sec:exp-search-strategies} of the paper, as the latter reports the mean and standard deviation.

We can draw three main conclusions from those plots. (i) On all instance size of the TSP, \compass clearly outperforms baselines, with a search that constantly find better solutions on average. (ii) We can clearly see the difference between principled search (\compass, EAS) and stochastic sampling (POMO, Poppy 16), as the former have an improving `latest batch performance', whereas the latter do not (iii) Interestingly, on several tasks, EAS has a higher maximum value on its latest batch (averaged on the 1000 problem instances); but the wider search of \compass enables to find better solutions for each problem instance in average. This can be observed on all JSSP sets and on the two first CVRP sets. For larger instances of CVRP, we can see that EAS is able to outperform our method: EAS is updating more parameters than \compass and this difference of modified parameters increases as the instance size increases (because its the product of the embedding size and the number of nodes in the instance). This larger number of updated parameters enables a better adaptation but also comes with a computational cost.

\section{Mutation Operators Used in the Generalization Tasks}
\label{sec:appendix-mutations}

\begin{figure}[h!]
  \centering
  \includegraphics[width=1.0\textwidth]{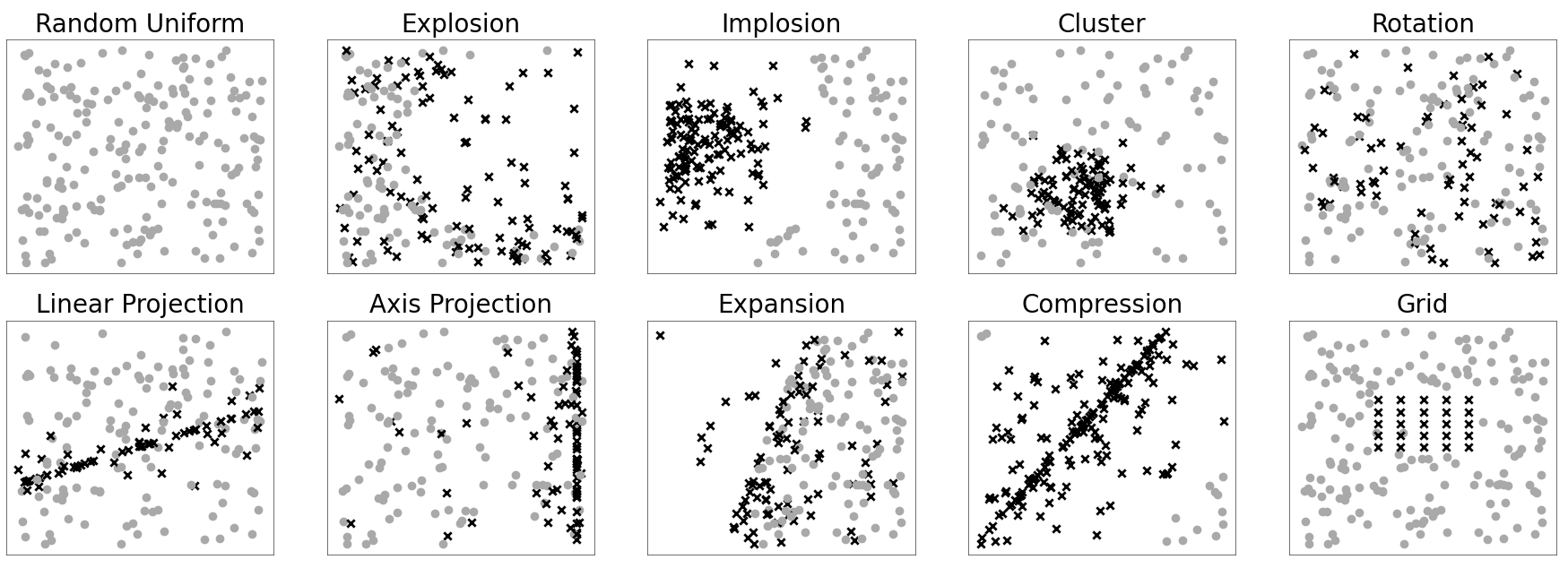}
  \caption{Visual representation of the RUE instance and mutated version of the instance through the 9 mutation operators. The gray dots represent the original (unmutated) cities, whereas, the black crosses represent the cities that have been mutated by the operator.}
  \label{fig:mutation-ops}
\end{figure}

In this section, we outline the mutation operators employed to generate the out-of-distribution (OOD) instances. The training distribution comprises random uniform Euclidean (RUE) instances, which are created by uniformly sampling the city coordinates from a unit square. To diversify the dataset and assess the methods' generalization abilities, these RUE instances can be mutated to exhibit different underlying distributions. We utilize nine mutation operators (taken from \citet{diverse_tsp_instances}) to construct the OOD dataset for TSP and CVRP by mutating the RUE instances. These mutated instances not only closely resemble real-world geographical data but also serve as valuable benchmarks for evaluating the methods' generalization capabilities. Figure~\ref{fig:mutation-ops} presents a visual representation of a RUE instance as well as the instance mutated by mutation operators and the list below defines each operator. 

\begin{itemize}
    \item Explosion: this operator simulates a random explosion that creates a gap or hole in the point cloud. It randomly selects a center of explosion, and then displaces all cities within a specified radius of this center to locations outside the radius.
    \item Implosion: this operator serves as the inverse of the explosion operator by bringing cities closer together towards a central point. It involves randomly selecting an implosion center and radius, and subsequently shifting all cities located within the implosion radius towards the center.
    \item Cluster: this operator generates a concentrated cluster of cities by randomly selecting a cluster center and mutating cities within a specified radius around the center. Specifically, cities are randomly chosen and their locations are modified to be within the selected radius of the cluster center.
    \item Rotation: this operator applies a rotation transformation to the cities around a specified pivot point. This mutation operator introduces angular displacement and rearranges the spatial arrangement of the cities in the TSP instance.
    \item Linear Projection: this operator performs a linear projection of the cities onto a randomly generated line. A random subset of cities is selected and repositioned along the line according to their original distances.
    \item Expansion: this operator merges the concepts of the explosion and linear projection mutations by displacing cities farther away from a randomly generated line in an orthogonal direction. 
    \item Compression: this operator, conversely to the expansion operator, is a combination of the implosion and linear projection operators by displacing the cities closer to the randomly generated line in an orthogonal direction. 
    \item Axis Projection: this operator is a special case of the linear project operator as the randomly generated line can either be parallel to the x- or y-axis.  
    \item Grid: this operator maps randomly chosen cities onto a grid-like structure. Specifically, the width, height, and proximity of the cities within the grid are randomly selected. Next, several cities are chosen from the instance and displaced into one of the grid locations. 
\end{itemize}

It is worth noting that all operators are parameterized by a probability argument which denotes the likelihood of mutating each city within the instance, referred to as mutation power. As a result, the mutation power is directly proportional to the number of cities mutated, and hence positively correlated with the \emph{shift} between the underlying distribution and the distribution of the RUE instance. Consequently, we use this mutation power as a reference to define the scale when studying the robustness of the baselines to out-of-distribution instances. 

\section{Architecture details}
\label{app:network-details}

\subsection{TSP and CVRP Networks}\label{app:tsp-crvp-network-details}

The architecture of our model for the TSP and CVRP problems consists of several key components. First, we have a single encoder responsible for encoding problem instances into a matrix of embeddings. This encoder follows a similar approach to other reinforcement learning methods for combinatorial optimization, such as POMO~\citep{POMO} and Poppy~\citep{poppy}.

Next, we have a single conditioned decoder that takes in the embeddings and the current state of the environment and outputs the next action. In contrast to the encoder being called at the beginning of the episode, the decoder is called at each step of the episode conditioned to the same latent vector throughout a given episode. The conditioned decoder architecture is quite similar to the one utilized in POMO and Poppy. It incorporates a multi-head attention mechanism to compute cross-attention between the embeddings and a local context. This local context includes the embedding of the starting point, the embedding of the current node, and the mean embedding of all nodes. The notable difference between our method \compass and prior works (POMO and Poppy) is that our decoder is conditioned on a vector sampled from a 16-dimension latent space. This latent vector is concatenated with the key, query, and value inputs of the multi-head attention decoder module. This conditioning allows us to create distinct policies while processing the same observation from the environment. Each latent vector corresponds to a unique policy, and thus, sampling the latent space to obtain vectors that our model can condition upon gives us an infinite set of policies.

\subsection{JSSP Network}
\label{sec:appendix-jssp-architecture}

The model architecture used for the JobShop Scheduling Problem (JSSP) is different from the networks used for TSP and CVRP (where the latter were taken from POMO \citep{POMO}). Prior work, which achieved state-of-the-art for JSSP \citep{hottung2022efficient}, use the L2D model \citep{Zhang2020} which is a Graph Neural Network, on a different, yet equivalent environment. We implemented an attention-based model and observed our model to outperform L2D and thus, decided to use our transformer-based architecture (similar to the TSP and CVRP models) to tackle JSSP. 

We utilize the actor-critic transformer architecture, as implemented in Jumanji \citep{jumanji2023github}. This architecture consists of an encoder and decoder network for both the actor and critic components. The encoder network incorporates attention layers for the machines' status, operation durations (with positional encoding), and joint sequence of jobs and machines. During each step of the episode, the encoder network is called and produces joint embeddings of the jobs and machines. These embeddings, along with the latent vector, are then fed into the decoder network. The decoder network receives the encoder embeddings and the latent vector as inputs and concatenates each dimension of the embedding with the latent vector before passing it through a multi-layer perceptron. At each step, the decoder network generates N marginal categorical distributions for each machine and provides a value generated by the critic. It is important to note that the actor and critic networks are separate entities with distinct sets of weights, ensuring that they do not share any parameters.

\section{Latent Policy Space}

\subsection{Design}
\label{sec:appendix-strategy-space}

The latent policy space defines a set of vectors that the single model can condition itself upon, and thus, this latent space provides us with an infinitely large set of policies (which becomes specialized and diverse through our training procedure). There are several ways to define the latent space, and the simplest approach is to use a set of $N$ one-hot encoded vectors. This way is very similar to the definition of the skill space in the RL Skill-Discovery literature~\citep{eysenbach2018diversity, sharma2019dynamics} and the difference between a set of independent policies and a single conditioned policy is reminiscent of the opposition between RL-based methods and Quality-Diversity methods for Skill Discovery~\citep{chalumeau2023neuroevolution}. Through preliminary experiments, we achieved a similar performance to Poppy 16~\citep{poppy} (current SOTA) with a set of 16 one-hot encoded vectors. However, the drawback of using a discrete latent space is that we cannot have an infinite set of policies and cannot interpolate within the latent space to adapt our model. Therefore, we define our latent space as a continuous n-dimension square. This enables us with an infinite number of policies that can be uniformly sampled during training and strategically searched during inference. During the experimentation phase, we investigated several different distributions and space sizes and concluded with a 16-dimensional space constrained to \([-1, 1]^{16}\). In practice, we multiply the latent vector by a factor 100, which is equivalent to sampling in \([-100, 100]^{16}\).

\subsection{Visualisation}
\label{sec:appendix-latent-space-plots}

\begin{figure}[h!]
  \centering
  \includegraphics[width=1.0\textwidth]{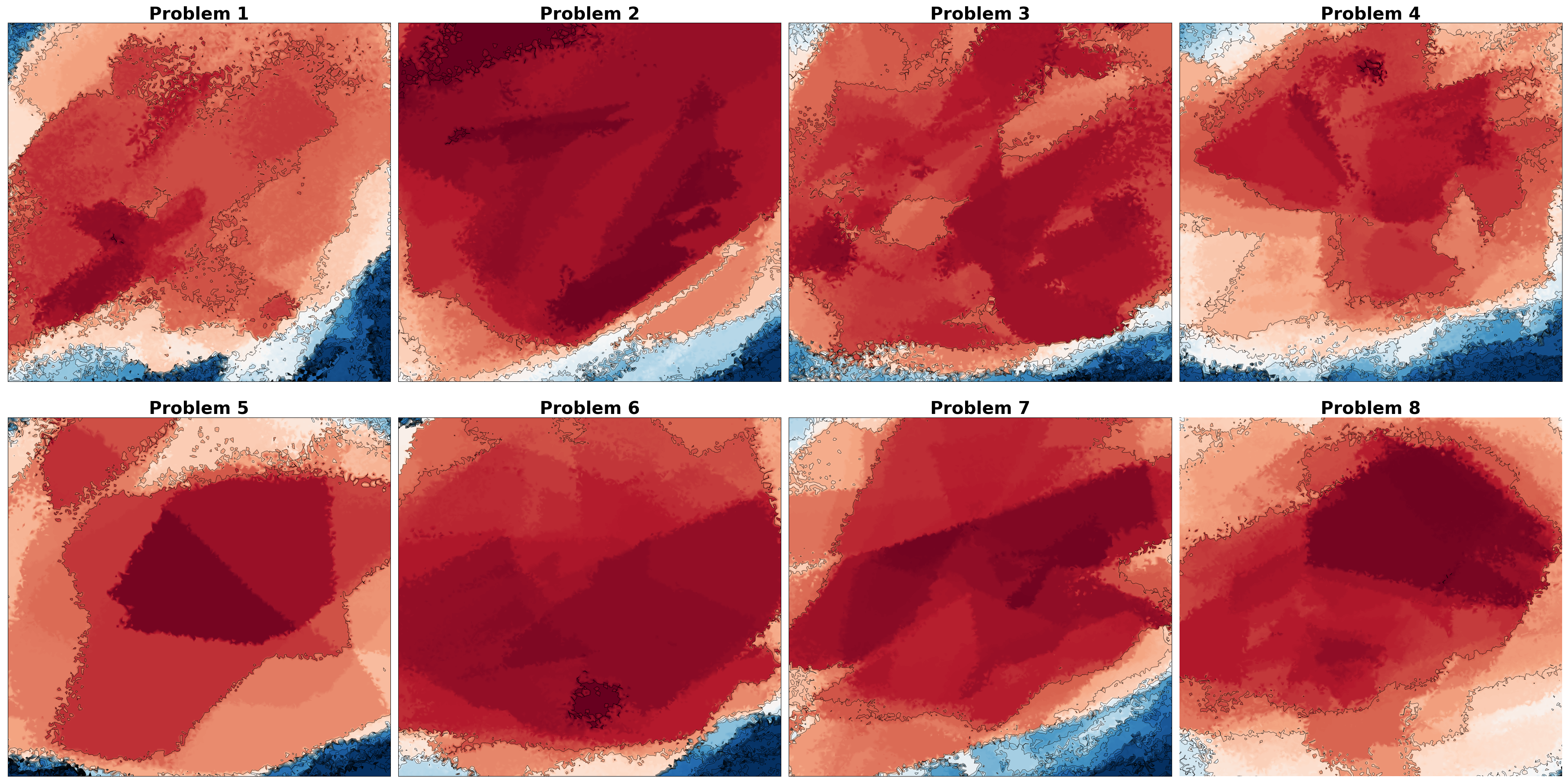}
  \caption{Latent space's heat map on 8 problem instances. We train a 2-dimensional \compass latent space and report its heat map on new unseen instances.}
  \label{fig:multiple_latent_space}
\end{figure}

In section~\ref{sec:exp-search-strategies}, we train \compass with a two-dimensional latent space and report the visualization of its performance landscape on a randomly selected instance of TSP\num{150}. To obtain this visualization, \compass was trained on a distribution of TSP\num{100}, and then we evaluated \num{32000} latent vectors on a randomly sampled TSP\num{150} instance. This enables us to create a precise heat-map (contour plot) of the performance landscape of the latent space on this instance. In Figure~\ref{fig:multiple_latent_space}, we report \num{7} additional visualizations for \num{7} new instances (the first one is the one reported in the main paper).

We can draw three main observations out of this plot: (i) the landscape is instance-dependent: we can see that the whole landscape changes and, in particular, the performant regions are different for each of the 8 problems studied. This shows that the whole latent space is used at inference time, and there does not seem to be a subpart dominating the others. (ii) They are usually several performant areas in the latent space for a given instance. This motivates the use of several independent search components at inference time. It enables us to avoid having all the search budget used on a suboptimal area. In case there is a clear area of interest, the components are likely to all converge there, hence having no loss of performance. This observation is insightful as it illustrates that several distinct solving-strategies can lead to solutions of similar quality (even if those solutions are distinct). (iii) Interestingly, they are clear discontinuities in the performance landscape. We can observe those in all problems visualized, but \emph{Problem 5} is a particularly relevant example. We can see several clear frontiers in the landscape, which shows that they can be a clear discontinuity in the mapping between solving-strategy and solution quality. We can make the hypothesis that there is one important decision that differs at the frontier, leading to a completely different performance.

Note that the deeper analysis of these latent spaces can provide very interesting insights about the solving process (e.g. the important decisions taken during the solving process), which could help to improve it (e.g. focusing the search on the important decision nodes). We leave this for future work.

\subsection{Exploring the Latent Policy Space at Inference Time}
\label{sec:appendix-ablation-search}

In this section, we present a comparison of various search methods applied to our latent policy space. Specifically, we analyze the performance of 4 strategies to illustrate choices made in the design of our method. In particular, we compare (1) a naive strategy, consisting of sampling 16 random vectors from the latent space, and sampling stochastically the resulting policies. This strategy is referred to as `fixed policies' because we are not re-sampling in the latent space. (2) a strategy consisting of sampling uniformly from the latent space and rolling out the resulting policy, referred to as `uniform-sampling'. This strategy can already illustrate the interest in having a latent space of diverse and specialized policies compared to sampling from a fixed set of policies. (3) CMA-ES search to navigate the latent space (4) our strategy, CMA-ES with several components (three), to illustrate the interest of being able to focus on distinct areas of the latent space.

\begin{figure}[h!]
  \centering
  \includegraphics[width=0.5\textwidth]{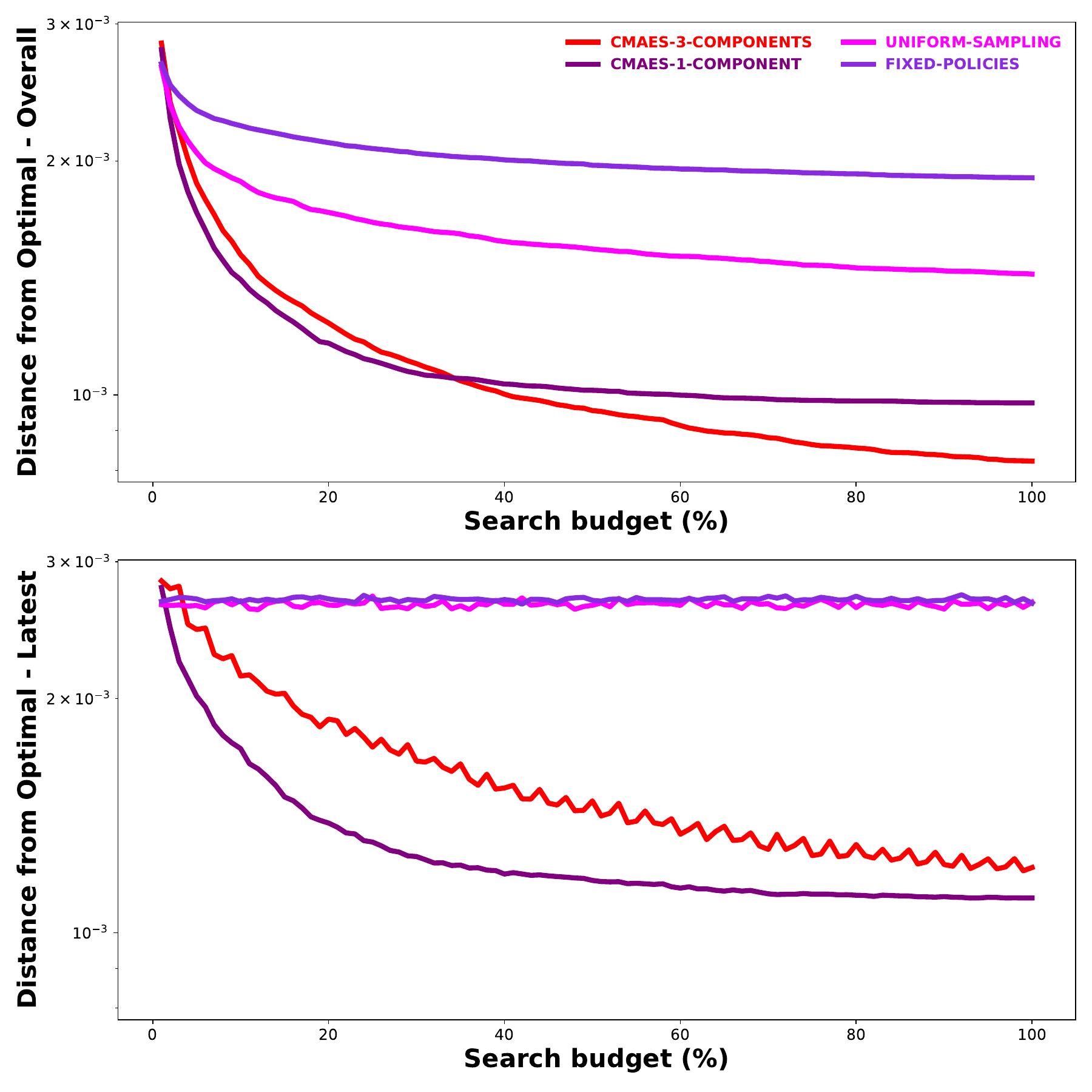}
  \caption{The performance of the different search strategies for TSP150 instances.}
  \label{fig:search-comparison}
\end{figure}

Figure~\ref{fig:search-comparison} illustrates the performances of the different search strategies on a set of TSP150 instances. We report the global performance as well as the latest batch performance. We can highlight three observations: (i) Sampling from the latent space brings significant improvement compared to sampling stochastically from a fixed set of policies. This shows the interest in having access to a space of diverse and specialized policies. (ii) Using an Evolution Strategy (like CMA-ES) to focus the search helps to make better use of the budget. We can see that the latest sampled solutions get better as the budget is used and the global performance improves faster compared to uniform sampling (iii) The final performance of the search gets better with three independent CMA-ES components rather than one. Being able to focus on several areas of the latent space enables us to avoid local optima and helps exploration. Interestingly, we can see that using all the budget for one component gives faster improvement but gets surpassed at the end of the budget. The choice of the accurate number of components is a middle ground between risking staying stuck in local optima and not having enough time to converge. In most of our tasks, using two or three components proved to work best.

\subsection{Alternative Strategies}
\label{sec:appendix-search-strategy}

At inference time, our goal is to explore the latent space to efficiently find promising areas of the latent space in order to obtain high-quality solutions within the given computational budget. As explained in Section~\ref{sec:method}, we decide to use multiple components of an evolution strategy to navigate our search space, because this approach is able to focus the search while being robust to local optima. Additionally, its parameters are easy to tune and work well on a wide range of tasks. Using multiple independent components of CMA-ES is also used in other works from the literature~\citep{fontaine2020cmame, cully2021multiemitter}, out of the CO scope. In practice, our implementation of CMA-ES is inspired by the GitHub package QDax~\citep{chalumeau2023qdax, lim2022accelerated}.

In Appendix~\ref{sec:appendix-ablation-search}, we provide a comparison of our search strategy (CMA-ES with multiple components) with alternatives. In particular, we compare CMA-ES with a single component to uniform sampling in the latent space. Nevertheless, we also considered other alternatives for this search. First, we explored Bayesian Search, as it is a data-efficient search that can hence be appropriate when the budget is limited, which is our case. Nevertheless, this approach was never able to do better than random sampling, although we tried multiple sets of hyper-parameters. A potential limitation of Bayesian Search is that it needs to model the space with a Gaussian Process, which can be very tedious if the space is noisy. Our main hypothesis is that the latent space of \compass is too noisy to be correctly modeled under the budget constraint. We see two mitigation of this effect. The first one is to add a regularization term during the training to create a `better-defined' latent space. The second would be to use a distribution to sample the space and use a Bayesian Search to optimize the parameters of this distribution. We leave these for future work.

Another alternative to Evolution Strategy is Gradient Descent, since we do have access to the derivatives. Nevertheless, gradient descent can easily be stuck in local optima, which is what we observe when analyzing the search of EAS in Section~\ref{sec:exp-search-strategies}. That being said, note that EAS and \compass could be combined, which could be expected to provide very good results, particularly in CVPR. We also defer this to future work.

\section{Training Procedure}
\label{sec:appendix-training-process}

In this section, we describe our model's training process. We first have a pre-training phase where we either reuse an existing model or train a single model with an encoder and a non-conditioned decoder using the REINFORCE algorithm.

Next, we begin the training procedure which aims to create a diverse set of specialized policies. In this process, we designate our set of policies with a 16-dimensional latent space that contains vectors that can be used to condition our decoder model (i.e., a policy is parameterized by the decoder model parameters and the latent vector). Specifically, the vector is concatenated with the key, query, and value inputs of the decoder model for each step of the episode. In order to use the learned decoder from the pre-training phase, we initialize the extra weights with zeros such that they do not initially affect the output. 

In the training procedure, the policy is trained to use the latent space to specialize to subareas of the problem distribution and this is achieved by solely training the policy conditioned on the latent variable that yields the highest reward for this instance. The details of the \compass training procedure is presented in Algorithm \ref{alg:compass} and can be understood as follows. At each iteration, we sample \(N\) vectors from the latent space \(\mathcal{Z}\) and a batch \(\mathcal{B}\) of instances from the problem distribution \(\mathcal{D}\). Then, for each instance \(\rho_i \text{ where } i \in 1, \dots, \mathcal{B}\) and sampled vector \(z_k \text{ where } k \in 1, \dots, N\), we rollout the conditioned policy \(\pi_\theta(\cdot | z_k)\) on the problem instance (i.e., generate a trajectory which represents a solution to the instance). Next, for each instance, we determine the best-performing latent vector and this is done by computing which conditioned policy obtained the highest reward on the instance. Finally, we only train the best latent vector for each problem instance and use the REINFORCE loss to perform backpropagation through the network parameters of our model (including both the encoder and decoder networks).

Note, we only train on instances that have a conditioned policy performing strictly better than the remaining policies. For example, if two policies have the exact same performance which is the maximum amongst the set of sampled policies, neither of the policies is trained on the instance. This approach enhances specialization and promotes a more balanced distribution of instances solved by each policy, resulting in a better spread of contributions within the latent space.

Lastly, it is also important to note that the number of vectors sampled during training (i.e., the number of conditioned policies competing for each instance) plays a significant role in the training process. The number of sampled latent vectors can be potentially infinite, the constraint lies in hardware or runtime limitations. Increasing the "number of samples $N$ in \compass promotes greater specialization and competition among policies in the latent space, which can be leveraged during inference when tackling new instances.

The final COMPASS neural solvers are trained until convergence, on a TPU v3-8. For each problem the training time and environment steps are: 4.5 days (110M steps) for TSP, 5.5 days (76.5M steps) for CVRP and 4.5 days (4.2M steps) for JSSP.

\begin{algorithm*}[htb]
	\caption{\compass Training}
	\label{alg:compass}
	\begin{algorithmic}[1]
		\STATE {\bfseries Input:} problem distribution \(\mathcal{D}\), latent space \(\mathcal{Z}\), number of samples $N$, batch size \(\mathcal{B}\), number of training steps $K$, policy $\pi_\theta$ with pre-trained parameters $\theta$. \;
            \STATE $\pi_\theta \leftarrow \FuncSty{Augment}(\pi_\theta)$ \; 
            \COMMENT{Augment the pre-trained policy to take as input the latent variable.}
		\FOR {step 1 to $K$}
			\STATE $\rho_i \leftarrow \FuncSty{Sample}(\mathcal{D}) ~ \forall i \in {1, \dots, \mathcal{B}}$ \;
                \STATE $z_i \leftarrow \FuncSty{Sample}(\mathcal{Z}) ~ \forall i \in {1, \dots, N}$
			\STATE ${\tau}^k_i \leftarrow \FuncSty{Rollout}(\rho_i, \pi_\theta(\cdot | z_k)) ~ \forall i \in {1, \dots, \mathcal{B}}, \forall k \in {1, \dots, N}$ \;
			\STATE $k^*_{i} \leftarrow \argmax_{k \leq N} \mathcal{R}({\tau}^k_{i}) ~ \forall i \in {1, \dots, \mathcal{B}}$ \; 
                \COMMENT{Select the best vector for each problem $\rho_i$.}
			\STATE $\nabla L(\theta) \leftarrow \frac{1}{B} \sum_{i \leq \mathcal{B}} \FuncSty{REINFORCE}({\tau}^{k^*_{i}}_{i})$  \; \COMMENT{\rep{Propagate the gradients}{Backpropagate} through these only.}
			\STATE $\theta \leftarrow \theta - \alpha \nabla L(\theta) $ \;
		\ENDFOR
	\end{algorithmic}
\end{algorithm*}

\section{Hyper-parameters}
\label{sec:appendix-hp}

In Table \ref{tab:hyperparams}, we report all the hyper-parameters of our method. Interestingly, our method is quite robust to these parameters and we almost use the same for all types of tasks. Note that for TSP sizes 125, 150, and 200, we report the hyper-parameters used during inference for the multi-components CMAES algorithm but there is no training hyper-parameters to report as we model used was trained on instances of size 100.

\begin{table}[h]
    \caption{The hyper-parameters used in \compass.}
    \centering
    \scalebox{0.95}{
        \begin{tabular}{c | c | c | c | c | c | c |}
        Phase & Hyper-parameters & TSP100 & TSP(125, 150) & TSP200 & CVRP & JSSP \\
        \midrule
         & latent space dimension & 16 & - & - & 16 & 16 \\
        Train time & training sample size & 128 & - & - & 128 & 128 \\
         & instances batch size & 8 & - & - & 8 & 8 \\
        \midrule
         & policy noise & 1 & 0.1 & 0.1 & 0.1 & 0.1 \\
        Inference & num. CMAES components & 3 & 3 & 2 & 2 & 3 \\
        Time & CMAES init. sigma & 100 & 100 & 100 & 100 & 100 \\
         & sampling batch size & 16 & 16 & 16 & 16 & 16 \\
        \end{tabular}
    }
    \label{tab:hyperparams}
\end{table}

\section{Model Checkpoints}
\label{sec:appendix-checkpoints}

We compare our method \compass to three main baselines: POMO \citep{POMO}, Poppy \citep{poppy}, and EAS \citep{hottung2022efficient} on three CO problem, TSP, CVRP, and JSSP. The checkpoints used to run the POMO and Poppy models for TSP and CVRP are taken from \citet{poppy}, and the EAS baseline is executed using the same POMO checkpoint. Those checkpoints are publicly available at \url{https://github.com/instadeepai/poppy}. For JSSP, we trained the single agent and Poppy models ourselves as they were not available. Lastly, we provide the \compass checkpoints used to obtain the results reported in this work. All those are available at \url{https://github.com/instadeepai/compass}.

\section{Performance with a Small Evaluation Budget}
\label{sec:appendix-fast-rl}

In Table \ref{fig:appendix_benchm_results}, we provide the performances of \compass and the baseline methods on the standard benchmark dataset with a smaller budget (typically 10\% of the usual budget). It can be seen that \compass outperforms the baselines on a majority of the tasks (specifically, 3/4 for TSP, 2/4 for CVRP, and all 3/3 for JSSP). In practice, a solver is particularly useful if it is performing well for a wide range of budget, in other words, able to provide good solutions fast, but also able to continually improve if any additional budget is given. \compass is able to provide an efficient to improve with a budget while being competitive (or outperforming) state-of-the-art methods with low budget.

\begin{table}[t]
  \centering
\caption{Results of \compass against the baseline algorithms for few-shot task on (a) TSP, (b) CVRP, and (c) JSSP problems. The methods are evaluated on instances from training distribution as well as on larger instance sizes to test generalization. The methods are only given 10\% of the standard budget.}
  \label{fig:appendix_benchm_results}
\begin{subtable}[b]{0.9\textwidth}
\caption{TSP}
  \centering
    \scalebox{0.8}{
    \begin{tabular}{l | cc | cc | cc | cc |}
        & \multicolumn{2}{c|}{\textbf{Training distr.}}
        & \multicolumn{6}{c|}{\textbf{Generalization}} \\
      & \multicolumn{2}{c|}{$n=100$} & \multicolumn{2}{c|}{$n=125$} & \multicolumn{2}{c|}{$n=150$} & \multicolumn{2}{c|}{$n=200$} \\
    Method & Obj. & Gap & Obj. & Gap & Obj. & Gap & Obj. & Gap \\
    \midrule
    Concorde & 7.765 & $0.000\%$ & 8.583 & $0.000\%$ & 9.346 & $0.000\%$ & 10.687 & $0.000\%$ \\
    LKH3 & 7.765 & $0.000\%$ & 8.583 & $0.000\%$ & 9.346 & $0.000\%$ & 10.687 & $0.000\%$ \\
    \midrule
    
    \begin{tabular}{@{}ll@{}}
    POMO (greedy)\\
    POMO\\
    \poppy 16 \\
    EAS \\
    \textbf{\compass (ours)} \\
    \end{tabular} &

    \begin{tabular}{@{}c@{}}
    7.796\\
    7.785\\
    7.769\\
    7.785\\
    \textbf{7.769} \\
    \end{tabular} &
    
    \begin{tabular}{@{}c@{}}
    0.404\% \\
    0.263\% \\
    0.045\% \\
    0.254\% \\
    \textbf{0.044\%} \\
    \end{tabular} &

    \begin{tabular}{@{}c@{}}
    8.635 \\
    8.619 \\
    \textbf{8.593} \\
    8.618\\
    8.595 \\
    \end{tabular} &
    
    \begin{tabular}{@{}c@{}}
    0.607\% \\
    0.424\% \\
    \textbf{0.115\%} \\
    0.403\% \\
    0.138\% \\
    \end{tabular} &

    \begin{tabular}{@{}c@{}}
    9.440 \\
    9.424 \\
    9.373\\
    9.419\\
    \textbf{9.373} \\
    \end{tabular} &
    
    \begin{tabular}{@{}c@{}}
    1.001\% \\
    0.835\% \\
    0.293\% \\
    0.776\% \\
    \textbf{0.283\%} \\
    \end{tabular} &

    \begin{tabular}{@{}c@{}}
    10.933 \\
    11.066 \\
    10.867 \\
    11.037 \\
    \textbf{10.787} \\
    \end{tabular} &
    
    \begin{tabular}{@{}c@{}}
    2.300\% \\
    3.545\% \\
    1.683\% \\
    3.276\% \\
    \textbf{0.933\%} \\
    \end{tabular}

    \end{tabular}}
\end{subtable}
\hfill
\begin{subtable}[b]{0.9\textwidth}
\caption{CVRP}
    \centering
    \scalebox{0.77}{
    \begin{tabular}{l | cc | cc | cc | cc |}
        & \multicolumn{2}{c|}{\textbf{Training distr.}}
        & \multicolumn{6}{c|}{\textbf{Generalization}} \\
      & \multicolumn{2}{c|}{$n=100$} & \multicolumn{2}{c|}{$n=125$} & \multicolumn{2}{c|}{$n=150$} & \multicolumn{2}{c|}{$n=200$} \\
    Method & Obj. & Gap & Obj. & Gap & Obj. & Gap & Obj. & Gap \\
    \midrule
    LKH3 & 15.65 & $0.000\%$ & 17.50 & $0.000\%$ & 19.22 & $0.000\%$ & 22.00 & $0.000\%$ \\
    \midrule
    
    \begin{tabular}{@{}ll@{}}
    POMO (greedy)\\
    POMO\\
    \poppy 32 \\
    EAS \\
    \textbf{\compass (ours)} \\
    \end{tabular} &

    \begin{tabular}{@{}c@{}}
    15.874\\
    15.767\\
    15.722\\
    15.743\\
    \textbf{15.681} \\
    \end{tabular} &
    
    \begin{tabular}{@{}c@{}}
    1.430\% \\
    0.75\% \\
    0.459\% \\
    0.592\% \\
    \textbf{0.201\%} \\
    \end{tabular} &

    \begin{tabular}{@{}c@{}}
    17.818 \\
    17.690 \\
    \textbf{17.633} \\
    17.646 \\
    17.648 \\
    \end{tabular} &
    
    \begin{tabular}{@{}c@{}}
    1.818\% \\
    1.085\% \\
    \textbf{0.758\%} \\
    0.834\% \\
    0.846\% \\
    \end{tabular} &

    \begin{tabular}{@{}c@{}}
    19.750 \\
    19.610 \\
    19.558 \\
    19.553 \\
    \textbf{19.532} \\
    \end{tabular} &

    \begin{tabular}{@{}c@{}}
    2.757\% \\
    2.031\% \\
    1.757\% \\
    1.730\% \\
    \textbf{1.626\%} \\
    \end{tabular} &

    \begin{tabular}{@{}c@{}}
    23.318\\
    23.817 \\
    23.907\\
    23.612 \\
    \textbf{22.946}\\
    \end{tabular} &
    
    \begin{tabular}{@{}c@{}}
    5.992\% \\
    8.258\% \\
    8.666\% \\
    7.329\% \\
    \textbf{4.298\%} \\
    \end{tabular}

    \end{tabular}}
\end{subtable}
\hfill
\begin{subtable}[b]{0.9\textwidth}
\caption{JSSP}
    \centering
    \scalebox{0.8}{
      \begin{tabular}{l | cc | cc | cc |}
        & \multicolumn{2}{c|}{\textbf{Training distr.}}
        & \multicolumn{4}{c|}{\textbf{Generalization}} \\
      & \multicolumn{2}{c|}{$10 \times 10$} & \multicolumn{2}{c|}{$15 \times 15$} & \multicolumn{2}{c|}{$20 \times 15$} \\
    Method & Obj. & Gap & Obj. & Gap & Obj. & Gap \\
    \midrule
    OR-Tools & 807.6 & $0.0\%$ & 1188.0 & $0.0\%$ & 1345.5 & $0.0\%$ \\
    \midrule

    \begin{tabular}{@{}ll@{}}
    L2D (Sampling) \\
    Single \\
    Poppy 16 \\
    EAS \\
    \textbf{\compass (ours)} \\
    \end{tabular} &
        \begin{tabular}{@{}c@{}}
    871.7 \\
    869.7 \\
    857.1 \\
    868.3 \\
    \textbf{851.9} \\
    \end{tabular} & 
        \begin{tabular}{@{}c@{}}
    $8.0\%$\\
    7.684\% \\
    6.126\% \\
    7.516\% \\
    \textbf{5.48\%} \\
    \end{tabular} & 
        \begin{tabular}{@{}c@{}}
    1378.3\\
    1312.4 \\
    1306.1\\
    1313.6\\
    \textbf{1301.1}\\
    \end{tabular} &
    
    \begin{tabular}{@{}c@{}}
    $16.0\%$\\
    10.47\% \\
    9.94\% \\
    10.572\% \\
    \textbf{9.518\%} \\
    \end{tabular} & 
    
    \begin{tabular}{@{}c@{}}
    1624.6\\
    1517.6\\
    1516.2\\
    1519.3\\
    \textbf{1504.6}\\
    \end{tabular} & 
    
    \begin{tabular}{@{}c@{}}
    $20.8\%$\\
    12.787\% \\
    12.686\% \\
    12.918\% \\
    \textbf{11.827\%} \\
    \end{tabular} \\

    \end{tabular}}
    \label{tab:jssp_size_full}
\end{subtable}
\end{table}

\section{Limitations}

In this section, we address three potential limitations of our method, \compass. First, our generalization capacity is directly linked to and potentially limited by the diversity created by specializing on the training distribution. In particular, our objective function does not explicitly incorporate any term to promote further diversity within the latent space. The specialization we observe during training is a result of training only the top-performing conditioned policy on each instance. Although this approach is straightforward and effective, it can be argued that adding an unsupervised term during training could lead to a more diverse and ultimately higher-performing set of policies. 

Another limitation lies in our training process: when sampling the latent space uniformly, the latent vectors evaluated may not include the true best vector, resulting in training a policy conditioned on a sub-optimal vector. This is both data-inefficient (evaluating several conditioned policies to only train on one trajectory) and sub-optimal (as specialization decreases by training a vector that does not correspond to the best conditioned policy).

To address this issue, we can consider two approaches. First, we can incorporate a search in the latent space during training to increase the likelihood of sampling the best vector. This way, we explore a wider range of latent vectors and improve the chances of discovering the optimal conditioning for each instance. Alternatively, we can train a prior distribution that, given an instance, provides information about the promising regions in the search space. By utilizing this prior distribution, we can sample latent vectors that are more likely to lead to better-performing conditioned policies. By leveraging prior knowledge or conducting a targeted search, we can enhance the efficiency and effectiveness of our training process.

Lastly, we believe there is potential for improving the efficiency of our search process with a better-defined latent space. We have observed that the latent space is noisy, which poses a challenge when employing a Bayesian search method for exploration. Moreover, we anticipate that our CMA-ES search could achieve faster convergence if the space exhibits smoother characteristics. One possible approach to address this is by incorporating a regularization term into our training objective, which would promote a smoother and more well-behaved latent space.

\section{Extended related work}
\label{appendix:extended-related-work}

We focus the literature review of the main paper (section \ref{related_work}) on construction methods trained with reinforcement learning. In this section, we present several improvement methods, i.e. methods that start with an existing solution and learn to directly modify this solution to create a new one; close to the concept of local search. Those are interesting alternatives to construction methods, although there limitation lies in that they are usually more problem-specific, and are also highly biased by the choice of the initial solution used.

NeuRewriter~\citep{chen2019learning} learns a policy that updates the solution, factorized in two steps: choosing the part of the solution, and then choosing the rule used to create a new solution. Note that this assumes the existence of a set of existing update rules to pick from. Concurrently, \citet{hottung2020neural} builds upon the Large Neighborhood Search framework, which consists of destroying and repairing solutions. This method relies on heuristics to destroy the solution and a neural policy cor the reconstruction mechanism, and this is learned end-to-end with reinforcement learning. Their approach is solely assessed on vehicle routing problems. \citet{kim2021learning} reduces the dependence on the initial solution by learning a constructive model that generates diverse solutions, called seeds, that are used as starting points for an improvement neural policy. \citet{Costa2020} learns a neural policy that selects 2-opt operators to improve solutions of the TSP, and \citet{wu2021learning} extends it to CVRP.

\citet{ma2021learning} improves the performance of improvement methods (in VRP) based on transformers neural models by ensuring that the embedding that is learned to encode the instance being solved better capture the structure of the vehicle routing problems.

Overall, once a first solution has been created, using improvement approaches can be seen as a concurrent approach to policy adaptation to make the best use of a given budget to reach the best possible solution.

Another related method that helps fine-tuning pre-trained checkpoints for larger scales tasks is Meta-SAGE~\citep{son2023metasage}. This method meta-learns how to scale embedding with respect to the instance sizes, improving performing a policy gradient fine-tuning. This method requires using a mixed-sized distribution at training time. 

\section{Analysis of time consumption in COMPASS}
\label{appendix:time-consumption}

We provide an analysis of the time consumption of \compass at inference time, by looking at its time performance on TSP100 on a TPU v3-8. 
The inference procedure can be decomposed into four steps: (1) encoding an instance (once per episode), decoding steps (99 times per episode), (3) environment steps (99 times per episode), and (4) CMA-ES steps (namely the sampling of the vectors and the update of the distribution parameters, once per episode). All values are estimated by averaging 500 evaluations and are reported in milliseconds (ms).

These results show that the adaptation mechanism of \compass (CMA-ES sampling and updates) is negligible compared to the remaining steps in the inference procedure, with a difference of three orders of magnitude. This explains the similarity in runtimes reported in~\cref{fig:benchm_results} between \compass, Poppy and POMO: the encoding and decoding are mostly identical, and the additional adaptation mechanism is not significant.

Interestingly, the encoding step is done only once for the whole budget. Hence, the bigger the budget, the more negligible the encoding steps become compared to the decoding and environment steps. Furthermore, the time difference between CMA-ES steps and all the other steps (aggregated over an episode) is only expected to grow with the size of the instance. First, because those individual steps depend on the instance size (larger matrix to encode or decode, more computations to be carried in the environment) whereas the CMA-ES adaptation mechanism does not depend on the instance size. Second, the number of the decoding and environment step increases with the instance size, which is not the case for the CMA-ES steps.

On the contrary, the adaptation mechanism used in EAS is time-consuming, making the method much slower than POMO, Poppy and \compass. Additionally, EAS' adaptation time increases as the instance size increases. 

\begin{table}[h!]
    \caption{The time taken in a full episode rollout of COMPASS for TSP100. All values are averaged over 500 evaluations. CMA-ES (sampling and update) takes three orders of magnitude less time than the decoding steps.} 
    \centering
    \scalebox{0.95}{
        \begin{tabular}{c | c | c | c | c |}
        Phase & Encoding & Decoding & Env. step & CMAES (sample \& update) \\
        \midrule
        Time for single event (ms) & 32.27 & 3.01 & 0.69 & 0.28 \\
        \midrule
        Occurrence in an episode & 1 & 99 & 99 & 1 \\
        \midrule
        \textbf{Time over an episode (ms)} & \textbf{32.27} & \textbf{297.99} & \textbf{68.31} & \textbf{0.28} \\
        \end{tabular}
    }
    \label{tab:time-consumption}
\end{table}

\section{Impacts of neural solver and search procedure in overall performance}

\subsection{Base solver vs adaptation mechanism}

Our method \compass has two important aspects: a conditioned neural solver, that enables to it capture specialised policies in a latent space; and an adaptation mechanism that searches the latent space at inference time. In this subsection, we provide more insight into the impact of both aspects on the overall performance observed.

Our experimental results provide evidence that both aspects – (1) a well-trained conditional neural solver and (2) an efficient search algorithm – are critical for strong performance.

Point (1) is illustrated by~\cref{fig:cmaes-search} and~\cref{fig:multiple_latent_space} which shows high-performing regions for a given instance. Point (2) is demonstrated by~\cref{fig:search-comparison} which shows the principled search method significantly outperforms random search. Additionally, we see that random search outperforms POMO and Poppy, confirming that the latent space “contains” high-performing and diverse policies.

\begin{figure}[h!]
    \centering
    \includegraphics[width=0.5\textwidth]{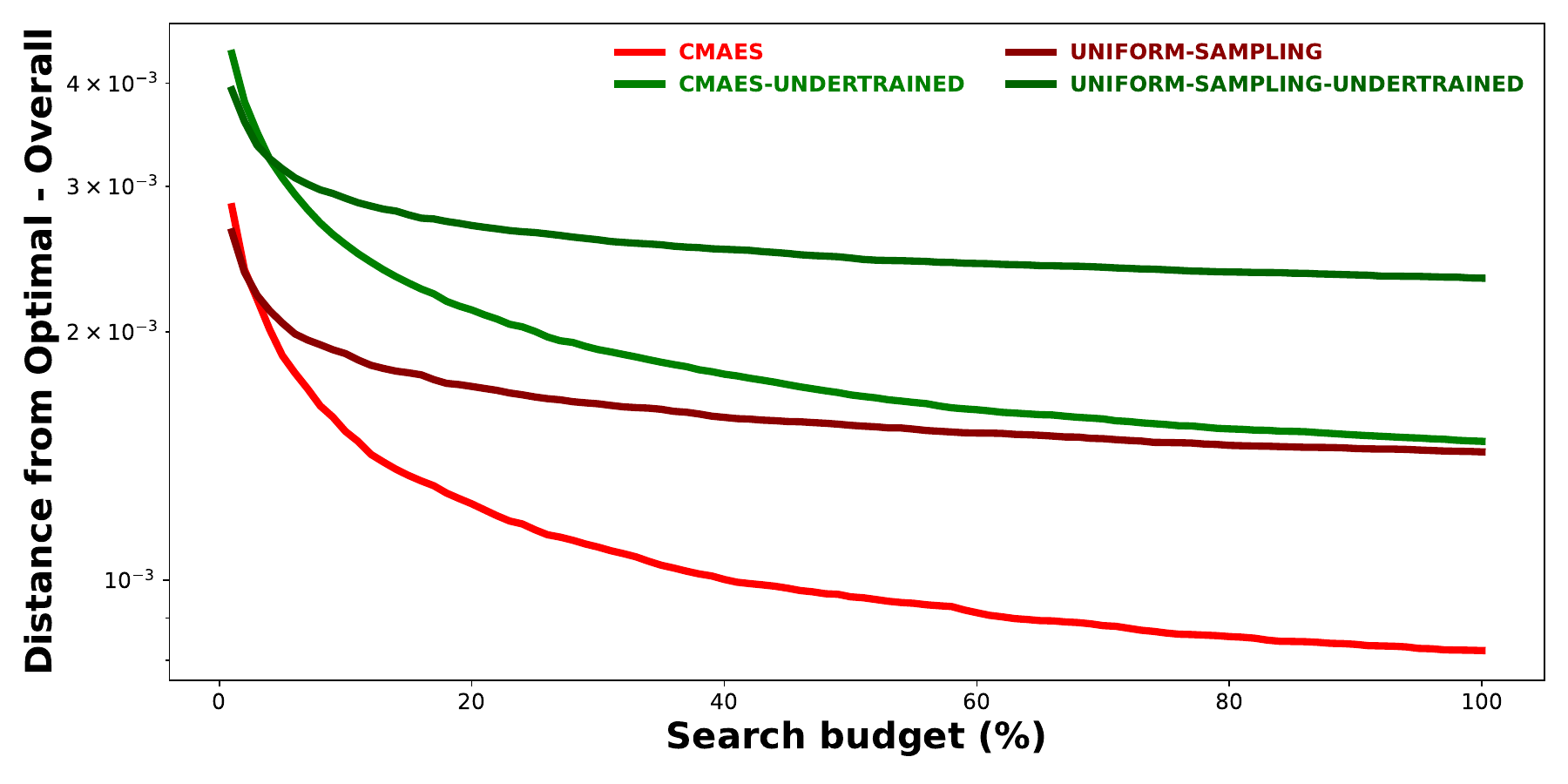}
    \caption{Performance of random and CMA-ES search on a conditioned solver that has not been trained until convergence (green) and a fully trained conditioned solver (red). Those are evaluated on 1000 instances of TSP150.}
    \label{fig:solver-undertrained}
\end{figure}

To further illustrate the importance of those combined aspects, we provide two additional experiments. First, we under-train a conditioned neural solver by stopping the training procedure well before convergence and compare two COMPASS models (fully- vs. under-trained) solving TSP150 instances with two search methods (CMA-ES and uniform sampling). Those results are reported in~\cref{fig:solver-undertrained}. The results demonstrate that: (i) both search methods for the fully trained model outperform those for the under-trained model, showing the importance of our training procedure. (ii) uniform search on the fully-trained solver outperforms CMA-ES search on the under-trained model, showing that the search alone is not sufficient.

\begin{figure}[h!]
    \centering
    \includegraphics[width=\textwidth]{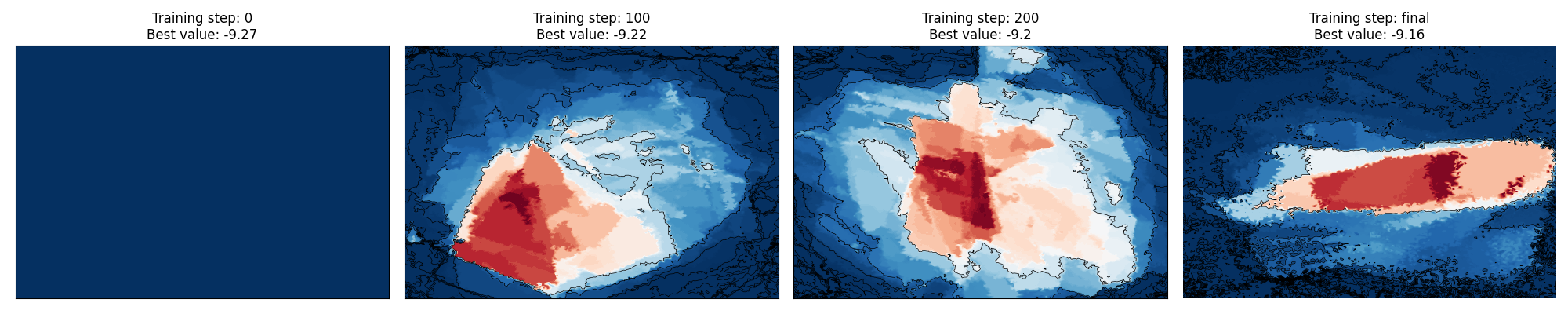}
    \caption{Evolution of the latent space during training on a given instance of TSP150.}
    \label{fig:latent-space-evolution}
\end{figure}

Second, we present the evolution of the latent space during training on a TSP150 instance, on~\cref{fig:latent-space-evolution}. It can be seen that initially, the space is uniform (no specialized regions exist). However, as training progresses, high-performing regions emerge (shown in red) which indicates the specialization of policies within the latent space, and we also see the improved performance of the best conditioned policy.

\subsection{Adaptation mechanism vs beam search}

Several strategies can be consider to optimally leverage a neural solver within the budget constraints to find the best possible solutions. To evaluate the efficiency and performance of our adaptation mechanism, CMA-ES search, we can compare to an alternative approach, such as Beam search. 

These two approaches are different paradigms: our approach searches the policy space and uses the policy to find a solution; Beam search explores directly the solution space, armed with a fixed policy. This can be approximated by comparing \compass with SGBS~\citep{choo2022simulationguided}, a heuristic approach to improve the performance of POMO~\citep{POMO}. A naive application of this heuristic on COMPASS (sampling a random policy with no latent space search) is equivalent to POMO+SGBS. We report the results of POMO+SGBS in~\cref{tab:additional_results} and show that \compass outperforms POMO+SGBS on the whole benchmark. This validates that it is worth searching for a good latent condition with the budget rather than fixing a random policy and using a beam search. Nevertheless, there may be a trade-off between search in latent space and heuristic solution search, which we leave for future work.

\section{Performance of our implementation of EAS}

In order to report the performance of EAS on our whole set of experiments and in similar conditions as POMO, Poppy and \compass, we have re-implemented EAS in our codebase, in Jax. This implementation is open-sourced\footnote{Implementations available at \url{https://github.com/instadeepai/compass}}. In this section, we report the results stated in the paper introducing EAS~\citep{hottung2022efficient} and report the results of our implementation in the same conditions.

Our implementation is faster on all TSP problem sets, and has similar speed on the CVRP sets. Our reported performance is better on all TSP sets and on half the CVRP sets. The difference observed on CVRP 150 and 200 cannot be certainly explained, it can be linked to difference between Jax and PyTorch or divergence in architectures that we could have missed. Note that our implementation is available online.

The results of EAS reported in this arxiv version is slightly different than the one reported in the OpenReview version because we have changed the implementation to better match the original one introduced in EAS~\citep{hottung2022efficient}. We used to backpropagate the gradients through the whole decoder to
update the embeddings instead of backpropagating only through the last attention layer, which was making the method significantly slower than expected. We have fixed this in our codebased and updated our results and paper accordingly in this arxiv version.

\begin{table}[h!]
    \centering
    \caption{Results of EAS (paper results vs. our implementation) with instance augmentation for (a) TSP and (b) CVRP.}
    \label{fig:eas_comparison}
    \begin{subtable}[b]{\textwidth}
    \caption{TSP}
      \centering
        \scalebox{0.65}{
        \begin{tabular}{l | ccc | ccc | ccc | ccc |}
            & \multicolumn{3}{c|}{\textbf{Training distr.}}
            & \multicolumn{9}{c|}{\textbf{Generalization}} \\
          & \multicolumn{3}{c|}{$n=100$} & \multicolumn{3}{c|}{$n=125$} & \multicolumn{3}{c|}{$n=150$} & \multicolumn{3}{c|}{$n=200$} \\
        Method & Obj. & Gap & {Time} & Obj. & Gap & {Time} & Obj. & Gap & {Time} & Obj. & Gap & {Time} \\
        \midrule
        
        \begin{tabular}{@{}ll@{}}
        EAS (paper) \\
        EAS (our implem.) \\
        \end{tabular} &
    
        \begin{tabular}{@{}c@{}}
        7.769\\
        7.768 \\
        \end{tabular} &
        
        \begin{tabular}{@{}c@{}}
        0.052\% \\
        0.038\% \\
        \end{tabular} &
    
        \begin{tabular}{@{}c@{}}
        5H \\
        3H \\
        \end{tabular} &
    
        \begin{tabular}{@{}c@{}}
        8.591\\
        8.590 \\
        \end{tabular} &
        
        \begin{tabular}{@{}c@{}}
        0.093\% \\
        0.080\%  \\
        \end{tabular} &
    
        \begin{tabular}{@{}c@{}}
        57M \\
        31M \\
        \end{tabular} &
    
        \begin{tabular}{@{}c@{}}
        9.363\\
        9.361 \\
        \end{tabular} &
        
        \begin{tabular}{@{}c@{}}
        0.182\% \\
        0.159\% \\
        \end{tabular} &
    
        \begin{tabular}{@{}c@{}}
        2H \\
        50M \\
        \end{tabular} &
    
        \begin{tabular}{@{}c@{}}
        10.730\\
        10.730 \\
        \end{tabular} &
        
        \begin{tabular}{@{}c@{}}
        0.402\%\\
        0.403\% \\
        \end{tabular} &
    
        \begin{tabular}{@{}c@{}}
        4H \\
        85M \\
        \end{tabular}
    \end{tabular}}
    \label{tab:eas_comparison_tsp}
    \end{subtable}
    \hfill
    \begin{subtable}[b]{\textwidth}
    \caption{CVRP}
        \centering
        \scalebox{0.65}{
        \begin{tabular}{l | ccc | ccc | ccc | ccc |}
            & \multicolumn{3}{c|}{\textbf{Training distr.}}
            & \multicolumn{9}{c|}{\textbf{Generalization}} \\
          & \multicolumn{3}{c|}{$n=100$} & \multicolumn{3}{c|}{$n=125$} & \multicolumn{3}{c|}{$n=150$} & \multicolumn{3}{c|}{$n=200$} \\
        Method & Obj. & Gap & {Time} & Obj. & Gap & {Time} & Obj. & Gap & {Time} & Obj. & Gap & {Time} \\
        \midrule
        
        \begin{tabular}{@{}ll@{}}
        EAS (paper) \\
        EAS (our implem.) \\
        \end{tabular} &
    
        \begin{tabular}{@{}c@{}}
        15.63\\
        15.62 \\
        \end{tabular} &
        
        \begin{tabular}{@{}c@{}}
        -0.13\% \\
        -0.175\% \\
        \end{tabular} &
    
        \begin{tabular}{@{}c@{}}
        9H \\
        7H \\
        \end{tabular} &
    
        \begin{tabular}{@{}c@{}}
        17.47 \\
        17.47 \\
        \end{tabular} &
        
        \begin{tabular}{@{}c@{}}
        -0.17\% \\
        -0.153\% \\
        \end{tabular} &
    
        \begin{tabular}{@{}c@{}}
        93M \\
        67M \\
        \end{tabular} &
    
        \begin{tabular}{@{}c@{}}
        19.22 \\
        19.26 \\
        \end{tabular} &
    
        \begin{tabular}{@{}c@{}}
        0.00\% \\
        0.213\% \\
        \end{tabular} &
    
        \begin{tabular}{@{}c@{}}
        3H \\
        108M \\
        \end{tabular} &
    
        \begin{tabular}{@{}c@{}}
        22.19 \\
        22.556 \\
        \end{tabular} &
        
        \begin{tabular}{@{}c@{}}
        0.86\% \\
        2.527\% \\
        \end{tabular} &
    
        \begin{tabular}{@{}c@{}}
        6H \\
        3H \\
        \end{tabular}
    
        \end{tabular}}
        \label{tab:eas_comparison_cvrp}
    \end{subtable}
\end{table}